\documentclass[11pt]{article}

\usepackage{arxiv}
\usepackage[utf8]{inputenc}
\usepackage[T1]{fontenc}
\usepackage{url}
\usepackage{subcaption}
\usepackage{booktabs}
\usepackage{amsfonts}
\usepackage{nicefrac}
\usepackage{microtype}
\usepackage{lipsum}
\usepackage{graphicx}
\usepackage{numprint}
\usepackage{titlesec}
\usepackage{XCharter}
\usepackage{amsmath}
\usepackage{adjustbox}
\usepackage{color, colortbl}
\usepackage{amssymb}
\usepackage{xcolor}
\usepackage{bbm}
\usepackage{datetime}
\usepackage[font = {footnotesize, it}]{caption}
\usepackage[ruled, vlined]{algorithm2e}
\usepackage{tabularx}
\usepackage{tablefootnote}
\usepackage{dsfont}
\usepackage{mathtools}
\usepackage{breakcites}
\usepackage{amsthm}
\usepackage{wrapfig}
\usepackage{float}
\usepackage[colorlinks = true, linkcolor = bluecite, urlcolor = bluecite]{hyperref}
\npthousandsep{,}

\DeclareMathOperator*{\argmin}{argmin}

\newcommand{\Esp}[1]{\mathbb{E}\! \left[ #1 \right]}

\newcommand{\defeq}{\vcentcolon=}
\newtheorem{Example}{Example}
\DeclareMathOperator*{\ave}{average} 

\definecolor{bluecite}{HTML}{0875b7}
\hypersetup{citecolor = bluecite}

\title{Enhancing Claim Classification with Feature Extraction from Anomaly-Detection-Derived Routine and Peculiarity Profiles}

\author{
    Francis Duval \\
    Chaire Co-operators en analyse des risques actuariels\\
    Département des mathématiques\\
    Université du Québec à Montréal\\
    Montréal, QC H2X 3Y7\\
    \texttt{duval.francis.2@courrier.uqam.ca} \\
    \And
    Jean-Philippe Boucher\\
    Chaire Co-operators en analyse des risques actuariels\\
    Département des mathématiques\\
    Université du Québec à Montréal\\
    Montréal, QC H2X 3Y7\\
    \texttt{boucher.jean-philippe@uqam.ca} \\
    \And
    Mathieu Pigeon\\
    Chaire Co-operators en analyse des risques actuariels\\
    Département des mathématiques\\
    Université du Québec à Montréal\\
    Montréal, QC H2X 3Y7\\
    \texttt{pigeon.mathieu.2@uqam.ca} \\
}

\begin{document}

\maketitle

\begin{abstract}
Usage-based insurance is becoming the new standard in vehicle insurance; it is therefore relevant to find efficient ways of using insureds' driving data. Applying anomaly detection to vehicles' trip summaries, we develop a method allowing to derive a ``routine'' and a ``peculiarity'' anomaly profile for each vehicle. To this end, anomaly detection algorithms are used to compute a routine and a peculiarity anomaly score for each trip a vehicle makes. The former measures the anomaly degree of the trip compared to the other trips made by the concerned vehicle, while the latter measures its anomaly degree compared to trips made by any vehicle. The resulting anomaly scores vectors are used as routine and peculiarity profiles. Features are then extracted from these profiles, for which we investigate the predictive power in the claim classification framework. Using real data, we find that features extracted from the vehicles' peculiarity profile improve classification.
\end{abstract}

\keywords{Automobile insurance \and Telematics car driving data \and Driving habits \and Unsupervised anomaly detection \and Supervised learning  \and Claim classification \and Feature extraction}

\section{Introduction and Motivations}\label{sec:intro}

Although the idea dates back to \cite{vickrey1968automobile}, usage-based insurance (UBI) is a fairly new insurance scheme mostly used in vehicle insurance, in which the insured's premium is estimated by making use of their driving data, which is recorded by a mechanism, usually an on-board diagnostics (OBD) device or a smartphone application. Nowadays, the latter is favored by insurers over the OBD device because it is cheaper and more practical. Determining a fair pure premium for each insured is a crucial task for any sensible insurance company; traditionally, automobile insurers have relied mainly upon static attributes related to the vehicle or the insured, which are indirectly related to accident risk. With the rise of telematics technology, it is now feasible for insurers to offer a more customized premium that is more in line with an insured's risk, which may now be determined by considering the insured's volume, habits and style of driving. UBI, or pricing using telematics data, seems likely to be the future standard in automobile insurance and the market share of UBI products is currently growing quickly. According to a study\footnote{\url{https://www.alliedmarketresearch.com/usage-based-insurance-market}} conducted by \textit{Allied Market Research}, a market research firm based in Portland, Oregon, the worldwide UBI market is expected to expand at a compound annual growth rate of $25.1$\% from $2020$ to $2027$, reaching US\$ $149.2$ billion by $2027$. The reason for this rapid expansion is that UBI products are valuable to both insurers and insureds. On one side, UBI benefits insurers because usage-based premiums tend to attract safer drivers\footnote{This is only true in a free market. In some regions, the UBI market is not free due to regulations that prevent insurers from applying a surcharge based on insureds' driving data.}, which lowers claim costs and increases profit margins. In other words, having more granular premiums allows them to avoid adverse selection and helps them stay competitive in a fierce market. On the other hand, it goes without saying that safer drivers and drivers with low driving volume are eager to buy this type of insurance product, which saves them money. UBI may even attract riskier drivers who wish to adjust their driving behavior in order to obtain a discount. Perhaps they are only risky drivers because they previously had few monetary incentives for safe driving. UBI products also seem to increase customer satisfaction because they allow customers to have more control over their premium. On top of that, usage-based premiums can help achieve societal goals because they promote less and safer driving, thereby reducing road congestion and pollution (see, for instance, \cite{litman2007} and \cite{bordoff2008}). They also help reduce discrimination by insurers based on immutable criteria, such as the gender of the insured. The COVID-19 pandemic has accelerated the shift toward telematics insurance. Indeed, with the recent boom in work-from-home culture, people are using their vehicles less and are therefore attracted to UBI, which is more beneficial to them.

For pricing purposes, insurers generally model claim risk using supervised learning models, typically generalized linear models (GLM; \cite{nelder1972generalized}, \cite{dionne1989generalization}), and use risk factors as independent variables. Traditionally, risk factors involved in pricing have been static insured- or vehicle-related features that are thought to be correlated with the policy risk, such as the insured's gender, the vehicle age, the bonus-malus level, etc. Such risk factors are easily used in prediction models because they can be encoded with just one entry per policy. With devices recording telematics data, each policy can now be linked with a complete history of driving information. One of the challenges insurers face is therefore to translate this data into relevant telematics-based risk factors. Perhaps the most obvious one that can be derived from driving data is the total distance driven during the policy, or mileage, which can be seen as a measure of risk exposure. Motor insurance in which pricing is done using mileage as a rate factor is often referred to as pay-as-you-drive (PAYD) insurance. It has been shown in several studies that distance driven is strongly linked with claim frequency (see for instance \cite{litman2005pay}, \cite{ferreira2010pay}, \cite{boucher2013pay} and \cite{lemaire2016use}). Consequently, some papers have focused upon this risk factor and have further investigated the relation between mileage and claim frequency: \cite{boucher2017exposure} make use of generalized additive models (GAM; \cite{hasti1990generalized}) and splines to model the non-linear effect of mileage on claim frequency and propose a pricing scheme that incorporates this information; more recently, \cite{boucher2020longitudinal} find, using generalized additive models for location, scale and shape (GAMLSS; \cite{rigby2005generalized}), that while the apparent association between mileage and claim frequency seems non-linear, the true relationship is quite linear. Accordingly, they conclude that each additional kilometer driven by an insured increases the expected number of claims by about $\frac{1}{\numprint{15000}}$, and state that this could be used to give a  surcharge/rebate to an insured that drives more/less than expected.

Although distance driven is probably the best telematics-based feature for predicting claims, it does not tell the whole story about an insured's risk and therefore, further driving data may be useful for pricing. Driving data can be divided into two classes: driving-habits-related data and driving-style-related data. The former includes information on \textit{when}, \textit{where} and \textit{how much} the insured drives, whereas the latter describes \textit{how} they drive. In this regard, motor insurance in which the premium is computed using driving style-related data is often referred to as pay-how-you-drive (PHYD) insurance. Some studies focus on driving habits-based risk factors (\cite{paefgen2013evaluation}, \cite{paefgen2014multivariate}, \cite{guillen2019use}). \cite{roel2017unraveling}, besides using mileage and number of trips, compute compositional predictors based on information such as the distribution of distance driven over different time slots and types of road. \cite{ayuso2019improving} propose a methodology in which driving habits-related risk factors are used as a correction to the premium calculated with traditional risk factors. Other studies, in conjunction with driving habits, investigate the predictive power of driving style-related data (\cite{wuthrich2017covariate}, \cite{gao2018driving}, \cite{gao2019claims}, \cite{gao2019convolutional}, \cite{huang2019automobile}, \cite{gao2021boosting}, \cite{so2021cost}). \cite{guillen2021near} develop a pricing scheme in which near-miss events, situations in which an accident is ``narrowly'' averted, and which comprise harsh braking, harsh acceleration and smartphone usage events, are used to update a baseline premium on a weekly basis.

Of the studies that investigate the relationship between driving data and claim risk, many use handcrafted telematics-based risk factors or, in other words, extract telematics features that are arbitrarily based on human guesswork. Besides total distance driven, some of the most popular handcrafted telematics features include the average distance per trip, the fraction of night and urban driving and the fraction of distance traveled above the speed limit (see, for instance, \cite{roel2017unraveling}, \cite{guillen2019use}, \cite{ayuso2016telematics}, \cite{huang2019automobile}, \cite{ayuso2019improving}, \cite{ayuso2014time}, \cite{ayuso2016using}, \cite{geyer2020asymmetric}). Although these handcrafted features often lead to a substantial improvement in claim risk modeling, they are arbitrary and subjective, which may favor or penalize some drivers. Take, for instance, the fraction of night driving. Where do you draw the line between night and day? If this line is at $00$:$00$, an insured who often drives between $11$:$30$ p.m. and $00$:$00$ will be favored, while another who often drives between $00$:$00$ and $00$:$30$ a.m. will be penalized, because night driving is often considered more dangerous than day driving. In this study, an automatic procedure that permits the extraction of telematics features from driving-habits-related data is developed. It has the benefit of not requiring human guesswork and is therefore more objective and fair. For this purpose, both a \textbf{routine} and a \textbf{peculiarity profile} are first derived for each vehicle by means of anomaly detection algorithms. Such algorithms allow the computation of an anomaly score, which measures the level of ``outlierness'' for each observation in a dataset. These algorithms are used to derive both a routine and a peculiarity score for each vehicle trip. For this, anomaly detection algorithms are applied using two different approaches, which are referred to as the \emph{local} and \emph{global schemes}. Scores calculated according to the former scheme correspond to the \emph{routine scores}, while those calculated with the latter scheme are referred to as the \emph{peculiarity scores}. In the local scheme, algorithms are applied to each vehicle individually and in silo, which means the anomaly score of a given trip made by a vehicle $x$ is defined only with respect to all the other trips made by vehicle $x$. Each driver thus defines their own normality against which their trips are compared. The routine-scores vector for a vehicle forms what we call its routine profile. In the global scheme, algorithms are applied to all trips by all vehicles at once, meaning that the anomaly score of a given trip made by vehicle $x$ is defined with respect to information on trips made by vehicle $x$ and all other vehicles. The peculiarity scores vector for a vehicle capture where its driving habits lie in relation to the other vehicles and is referred to as its peculiarity profile. We believe that the insureds' routine and peculiarity profiles can help differentiate between claimants and non-claimants. Our hypothesis is that more routine drivers, who know more about the trips they make, are less prone to claim than less routine ones. Indeed, an insured who always makes the same trips knows their way and how to cope with sensitive parts of it. On the other hand, we believe drivers who tend to make peculiar trips are more likely to claim than others because it is expected that peculiar trips, which are made, for instance, at unusual times of the day or speeds, are more dangerous. Features are subsequently extracted from these profiles, whereupon their predictive power is investigated through an elastic-net logistic regression model. In particular, we answer the question of how an insured's routine and peculiarity profile influences their likelihood of claiming, which has never been addressed before in the literature.

In addition to a traditional automobile insurance pricing dataset, we have at hand information on the driving habits of the vehicle's main driver in the format of trip summaries. Both datasets are described in Section~\ref{sec:data}. Before being fed to the anomaly detection algorithms, telematics data need to be preprocessed, which is done in Section~\ref{sec:Telematics data preprocessing and visualization}, together with some data visualization. Section~\ref{sec:Anomaly Detection Algorithms} follows, where the three anomaly detection algorithms used, namely Mahalanobis' method, the Local Outlier Factor and the Isolation Forest, are presented in detail. The claim classification model we use, which is a logistic regression model with elastic-net regularization, is then described in Section~\ref{sec:Supervised Binary Classification Model}. We also detail all the preprocessing steps required for the input features. Both the anomaly detection algorithms and the classification model have hyperparameters that need to be calibrated, which is done in Section~\ref{sec:Hyperparameter tuning}. In Section~\ref{sec:Analyses}, models using features extracted from the routine and peculiarity profiles are scored on the testing set, whereupon their performance is compared to a baseline model. We conclude in Section~\ref{sec:Conclusions}.

\section{Data Description}\label{sec:data}

\subsection{Telematics dataset}

The data we have at hand is provided by a Canadian property and casualty insurer. We have access to a telematics dataset in which each entry matches a trip made with an insured vehicle. The recording of a trip, which is done using an OBD device, starts when the engine is turned on and ends when it is turned off. This type of recording is sometimes referred to as a \textit{key-on key-off} event (see for instance \cite{roel2017unraveling}). The dataset is comprised of $\numprint{7438883}$ of these events reported by $\numprint{4834}$ vehicles, each of which is identifiable by its vehicle identification number (VIN). All of them were observed over a one-year period coinciding with a one-year insurance policy, starting somewhere between December $30^\text{th}$, $2015$ and January $1^\text{st}$, $2019$. Each trip is characterized by four quantities: the datetime of departure and arrival, the distance driven and the maximum speed reached. Along with the VIN and an identifier for trips, this translates into a six-column dataset, an extract of which is shown in Table~\ref{tab:1}.
\begin{table}[ht]
    \centering
    \begin{adjustbox}{max width = \textwidth}
        \begin{tabular}{c c c c c c}
            \toprule 
            \textbf{VIN} & \textbf{Trip ID} & \textbf{Departure datetime} & \textbf{Arrival datetime} & \textbf{Distance} & \textbf{Maximum speed}\\ 
            \midrule
            A & $1$ & $2017$-$05$-$02$ {$19$:$04$:$15$} & $2017$-$05$-$02$ {$19$:$24$:$24$} & $25.0$ & $104$\\
            A & $2$ & $2017$-$05$-$02$ {$21$:$31$:$29$} & $2017$-$05$-$02$ {$21$:$31$:$29$} & $6.4$ & $66$\\
            \vdots & \vdots & \vdots & \vdots & \vdots & \vdots \\
            A & $2320$ & $2018$-$04$-$30$ {$21$:$17$:$22$} & $2018$-$04$-$30$ {$21$:$18$:$44$} & $0.2$ & $27$\\
            \cmidrule(l){1-6}
            B & $1$ & $2017$-$03$-$26$ {$11$:$46$:$07$} & $2017$-$03$-$26$ {$11$:$53$:$29$} & $1.5$ & $76$\\
            B & $2$ & $2017$-$03$-$26$ {$15$:$18$:$23$} & $2017$-$03$-$26$ {$15$:$51$:$46$} & $35.1$ & $119$\\
            \vdots & \vdots & \vdots & \vdots & \vdots & \vdots \\
            B & $1485$ & $2018$-$03$-$23$ {$20$:$07$:$08$} & $2018$-$03$-$23$ {$20$:$20$:$30$} & $10.1$ & $92$\\
            \cmidrule(l){1-6}
            C & $1$ & $2017$-$11$-$20$ {$08$:$14$:$34$} & $2017$-$11$-$20$ {$08$:$40$:$21$} & $9.7$ & $78$ \\
            \vdots & \vdots & \vdots & \vdots & \vdots & \vdots \\
            \bottomrule 
        \end{tabular}
    \end{adjustbox}
    \caption{Extract from the telematics dataset. Dates are displayed in the yyyy-mm-dd format. The actual VINs have been hidden for privacy purposes.} 
    \label{tab:1}
\end{table}

\subsection{Traditional dataset}

We also have a $\numprint{4834}$-row dataset that provides us with information about the policy during the one-year observation period for each of the $\numprint{4834}$ vehicles in the telematics dataset. Each entry in this dataset depicts an insurance policy with ten risk factors that are traditionally leveraged in automobile insurance pricing, related to either the policyholder or the insured vehicle, all of which are detailed in Table~\ref{tab:classic}.
\begin{table}[ht]
    \centering
    \begin{adjustbox}{max width = \textwidth}
        \begin{tabular}{l l l}
            \toprule 
            \textbf{Variable name} & \textbf{Description} & \textbf{Type} \\
            \midrule
            \texttt{vin} & Unique vehicle identifier & ID\\
            \cmidrule(l){1-3}
            \texttt{annual\_distance} & Annual distance declared by the insured & Numeric\\
            \texttt{commute\_distance} & Distance to the place of work declared by the insured & Numeric\\
            \texttt{conv\_count\_3\_yrs\_minor} & Number of minor contraventions in the last three years & Numeric\\
            \texttt{gender} & Gender of the insured & Categorical\\
            \texttt{marital\_status} & Marital status of the insured & Categorical\\
            \texttt{pmt\_plan} & Payment plan chosen by the insured & Categorical\\
            \texttt{veh\_age} & Vehicle age & Numeric\\
            \texttt{veh\_use} & Use of the vehicle & Categorical\\
            \texttt{years\_claim\_free} & Number of years since last claim & Numeric\\
            \texttt{years\_licensed} & Number of years since obtaining driver's license & Numeric\\
            \cmidrule(l){1-3}
            \texttt{claim\_ind} & Indicator of the occurrence of a claim & Categorical\\
            \bottomrule 
        \end{tabular}
    \end{adjustbox}
    \caption{Overview of the traditional dataset.} 
    \label{tab:classic} 
\end{table}
This dataset, which we will refer to as the traditional dataset, also features the target variable that will later be used for claim classification, namely the indicator of the occurrence of a claim during the coverage period. Finally, there is the VIN column, which we need to merge the two datasets. Our goal is to extend the traditional dataset with features extracted from the telematics dataset.

\subsection{Training and testing sets}

In order to properly assess the performance of the models on new instances, a portion of the data must be set aside. To this end, $70$\% of the VINs (that is, $3384$ VINs) are used to build the training set, whereas the remaining $30$\% (that is, $1450$ VINs) comprise the testing set. The training set will later be used to tune and train the models, while the testing set will only come into play at the very end of the modeling process to assess the generalization performance of the previously tuned and trained models.

\section{Telematics data preprocessing and visualization}\label{sec:Telematics data preprocessing and visualization}

\begin{figure}
  \centering
  \subcaptionbox{Non-cyclical encoding\label{fig:sub-first}}
    {\includegraphics[width = \textwidth]{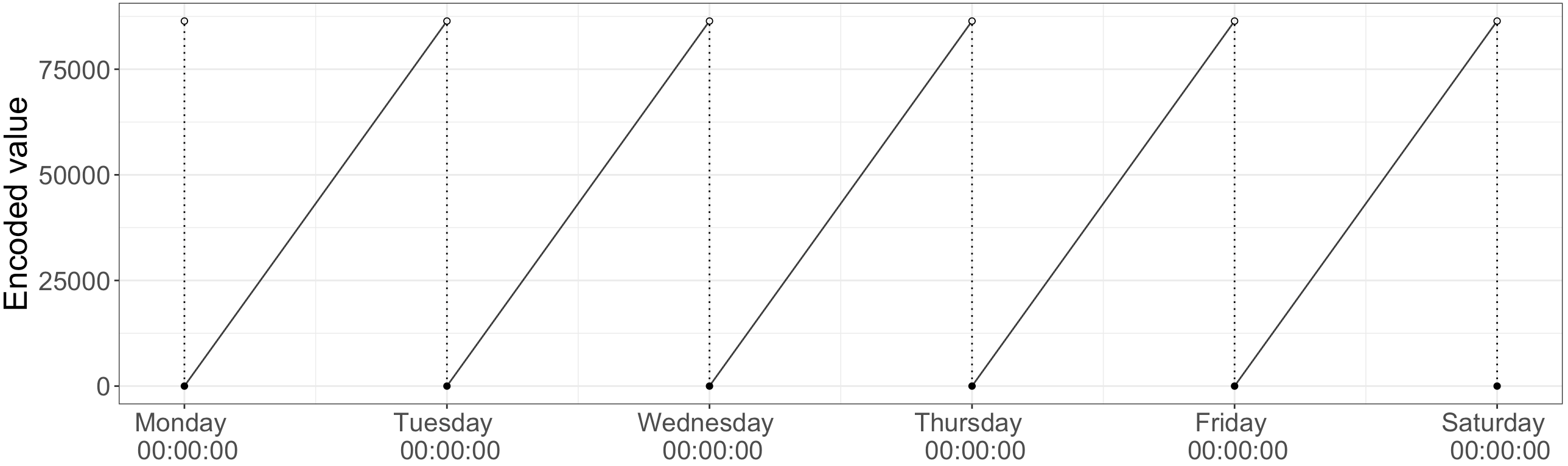}}
  \subcaptionbox{Sine and cosine encodings\label{fig:sub-second}}
    {\includegraphics[width = \textwidth]{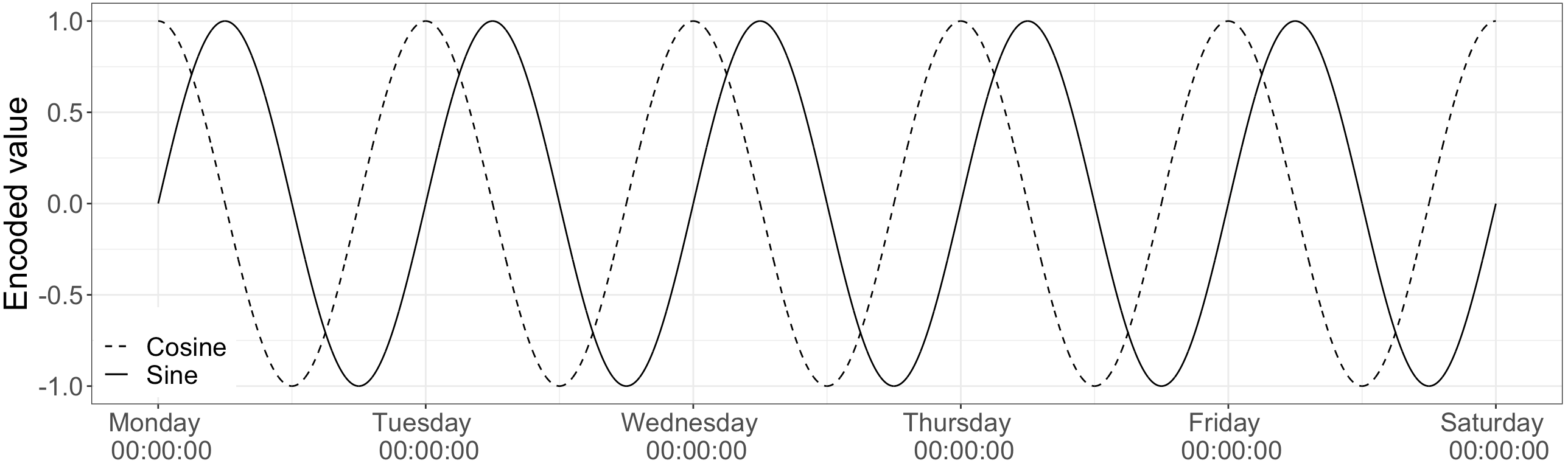}}
  \caption{Encoding of the time-of-day attribute (number of seconds elaspsed since midnight)}\label{fig:fig}
\end{figure}

\begin{figure}
    \centering
    \includegraphics[width = \textwidth]{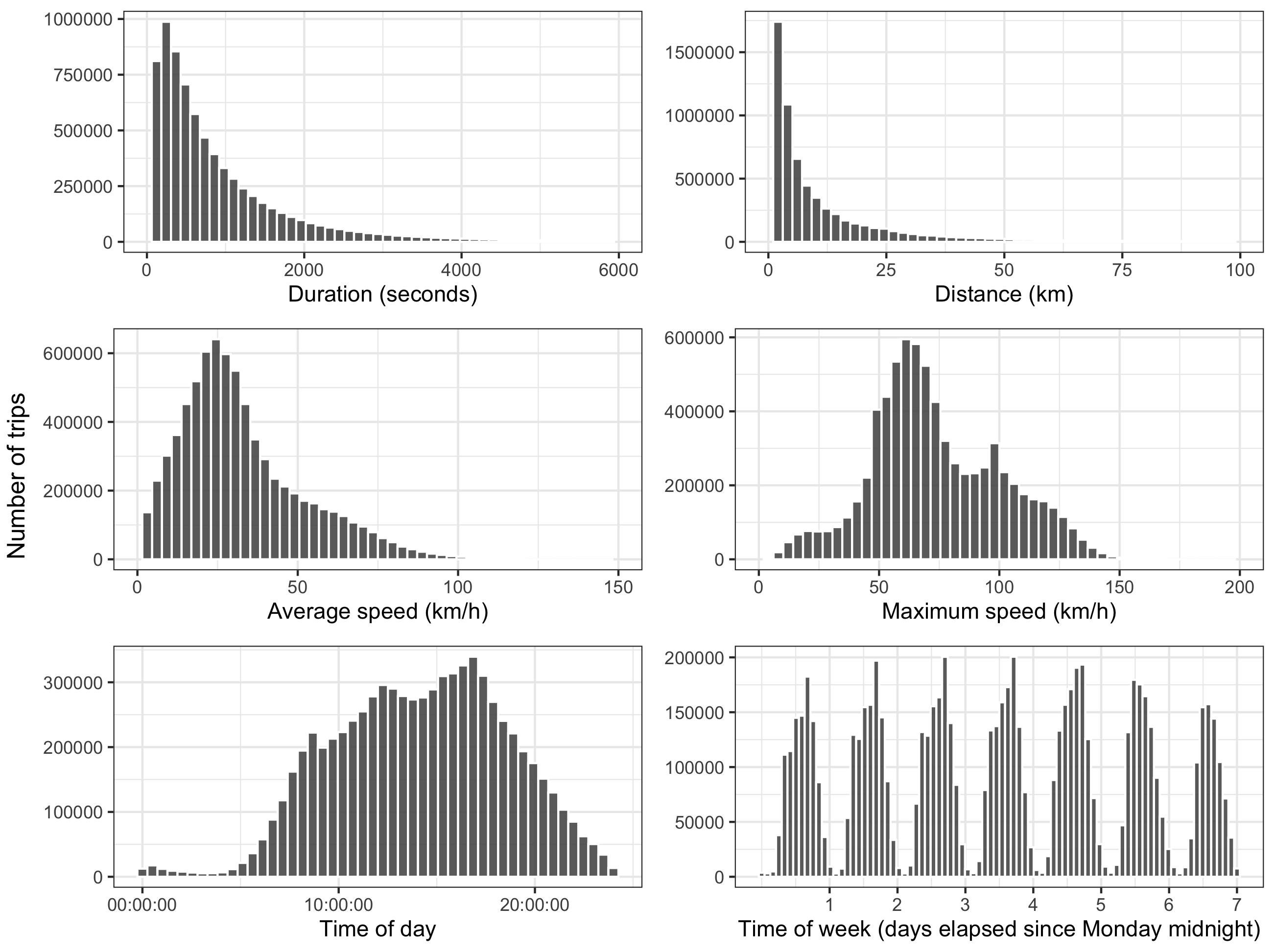}
    \caption{Distribution of the trip attributes.}
    \label{fig:telematics_histograms}
\end{figure}

In Section~\ref{sec:Anomaly Detection Algorithms}, three popular anomaly detection (AD) algorithms are introduced, each of which is used to compute both a routine and a peculiarity score for each vehicle trip. Note that since the routine scores are calculated in the local scheme, these are also referred to as \emph{local anomaly scores}, while peculiarity scores, which are calculated through the global scheme, are also called \emph{global anomaly scores}. Both local and global anomaly scores are calculated exclusively from information found in the telematics dataset of Table~\ref{tab:1}. However, we first preprocess the latter in order to extract useful information. Indeed, the raw telematics data we have are not in the ideal format to use as input to the AD algorithms. First of all, the departure datetime, which is stored as the time (in seconds) elapsed since an arbitrary point in time (January $1^\textbf{st}$, $1970$ in the \texttt{R} software), is not very meaningful for AD algorithms. It might be more useful to have information about the time of day and the day of the week the trip started. From the departure datetime, we thus create two new trip attributes, namely the time elapsed since midnight (in seconds) and the time elapsed since Monday midnight (in days), respectively referred to as the time-of-day and time-of-week trip attributes. These two newly created variables are cyclical in nature and need to be encoded accordingly. Let us consider the time-of-day attribute, which ranges from $0$ to $\numprint{86400}$. The non-cyclical encoding of this attribute over five days as a function of the datetime is shown in Figure~\ref{fig:sub-first}. Looking at the latter, one can understand the concern with supplying non-properly encoded cyclical data into the algorithms: there are discontinuities in the graph at the end of each day (when time goes from $23$:$59$:$59$ to $00$:$00$:$00$). Indeed, $00$:$00$:$00$ is encoded as $0$ whereas $23$:$59$:$59$ is encoded as $\numprint{86399}$. This create inconsistencies: without cyclical encoding, a trip starting at $23$:$59$:$59$ would be considered very ``far'' from another trip that starts at $00$:$00$:$00$ according to the AD algorithms. This is obviously an issue because the two trips are only one second apart. In order to address this, we use sine and cosine functions, which are cyclical, to encode both time-of-day and time-of-week attributes. To this end, we first normalize both of them so that a cycle (which is $\numprint{86400}$ seconds for the time-of-day attribute and seven days for the time-of-week attribute) is compressed between $0$ and $2\pi$, and then apply the sine/cosine function. Mathematically, the sine and cosine transformations of the time-of-day attribute are expressed with
\footnotesize
\begin{align*}
    \text{time\_of\_day}_{sin} = \sin\left(\frac{2\pi}{86400}\times\text{time\_of\_day}\right),\\
    \text{time\_of\_day}_{cos} = \cos\left(\frac{2\pi}{86400}\times\text{time\_of\_day}\right).
\end{align*}
\normalsize
In a similar fashion, the time-of-week attribute is encoded with
\footnotesize
\begin{align*}
    \text{time\_of\_week}_{sin} = \sin\left(\frac{2\pi}{7}\times\text{time\_of\_week}\right),\\
    \text{time\_of\_week}_{cos} = \cos\left(\frac{2\pi}{7}\times\text{time\_of\_week}\right).
\end{align*}
\normalsize
Sine and cosine encodings for the time-of-day attribute are shown in Figure~\ref{fig:sub-second}, where one sees that discontinuities have disappeared, which is desirable.
It is worth noting that these cyclical attributes need to be encoded with two dimensions, that is, with both sine and cosine functions. Indeed, if they are encoded with only the sine (or cosine) transform, the encoding would not be unique. For instance, in the case of the time-of-day attribute, the sine transform for midnight ($\numprint{0}$ seconds since midnight) is the same as for noon ($\numprint{43200}$ seconds since midnight), since 
\begin{align*}
    \sin\left(\frac{2\pi}{86400}\times0\right) = \sin\left(\frac{2\pi}{86400}\times43200\right) = 0.
\end{align*}
Secondly, two additional trip attributes are created from the four existing ones in Table~\ref{tab:1}, namely the average speed (in kilometers per hour) and the duration (in minutes) of the trip. Finally, because anomaly detection algorithms benefit from centered and scaled data, trip attributes undergo a z-score normalization. Consider the numerical vector $\boldsymbol{x} = (x_1, \dots, x_n)$. The z-score normalized version of $\boldsymbol{x}$ is
\begin{align*}
    \boldsymbol{x}^* = \left(\frac{x_1 - \overline{x}}{s}, \dots, \frac{x_n - \overline{x}}{s}\right),
\end{align*}
where $\overline{x} = \frac{1}{n}\sum_{i = 1}^n x_i$ and $s = \sqrt{\frac{1}{n-1} \sum_{i = 1}^n (x_i - \overline{x})^2}$. In essence, AD algorithms are applied on the preprocessed version of the telematics dataset of Table~\ref{tab:1}, for which the eight attributes are described in Table~\ref{tab:telematics}.
\begin{table}[ht]
    \centering
    \begin{adjustbox}{max width = \textwidth}
        \begin{tabular}{l l}
            \toprule 
            \textbf{Trip attribute} & \textbf{Description}\\
            \midrule
            \texttt{duration} & Trip duration (in minutes)\\
            \texttt{distance} & Trip distance (in kilometers)\\
            \texttt{avg\_speed} & Trip average speed (in kilometers per hour)\\
            \texttt{max\_speed} & Trip maximum speed (in kilometers per hour)\\
            \texttt{time\_of\_day\_sin} & Sine encoding of the number of seconds elapsed since midnight\\
            \texttt{time\_of\_day\_cos} & Cosine encoding of the number of seconds elapsed since midnight\\
            \texttt{time\_of\_week\_sin} & Sine encoding of the number of days elapsed since Monday midnight\\
            \texttt{time\_of\_week\_cos} & Cosine encoding of the number of days elapsed since Monday midnight\\
            \bottomrule 
        \end{tabular}
    \end{adjustbox}
    \caption{Variables (or trip attributes) of the eight-dimensional dataset to which anomaly detection algorithms are applied. All eight variables undergo a z-score normalization.} 
    \label{tab:telematics} 
\end{table}
Because the AD algorithms are applied to such a dataset, the anomaly degree of a trip is defined exclusively in terms of these attributes, which may be summarized as the driving habits. Therefore, AD algorithms seek to identify trips that exhibit an anomalous combination of duration, distance, average speed, maximum speed, time of day and time of week: the more anomalous the combination of attributes, the higher the anomaly score. In Figure~\ref{fig:telematics_histograms}, the distributions of the eight attributes of Table~\ref{tab:telematics} over the whole telematics dataset are shown. In fact, only six distributions are plotted because for both the ``time of day'' and ``time of week'' attributes, the non-encoded version is shown to facilitate interpretation.
As can be seen, trips of more than $50$ km are fairly rare, as are night trips. The vast majority of trips are made at an average speed of less than $100$ km/h and at a maximum speed of less than $150$ km/h. Moreover, the insureds in our portfolio drive slightly less on Sunday than on any other day. It should be noted that all trips were made in an area where the maximum speed limit is $110$ km/h.

\section{Multivariate Anomaly Detection Algorithms}\label{sec:Anomaly Detection Algorithms}

In this section, three popular anomaly detection algorithms are presented, each of which is used to derive both a routine and a peculiarity profile for each vehicle or, equivalently, both a local and a global anomaly score for each vehicle trips. Because the vehicles' trips are not labeled as anomalies or non-anomalies, we find ourselves more specifically in the framework of \emph{non-supervised} anomaly detection. Therefore, the three algorithms used are selected from non-supervised anomaly detection algorithms, which do not require a label to derive anomaly scores.

In the local scheme, the anomaly score of a trip, which we refer to as ``local,'' is computed by applying an AD algorithm to each vehicle separately, and thus reflects its anomaly degree with respect to all the other trips made by the vehicle in question. This means that a trip that seems very unlikely, such as a $200$ kilometer trip made on Tuesday at $3$ a.m., could be assigned a low local anomaly score, as long as the vehicle regularly takes this kind of trip. On the other hand, a trip that seems very common, for instance a ten kilometer trip on Monday at rush hour, could be assigned a high local anomaly score if the vehicle rarely makes this kind of trip. A trip with a high local anomaly score is one that bears little resemblance to trips typically made by the vehicle. The set of local anomaly scores for a vehicle's trips forms a vector of real numbers, or empirical distribution, which we call the ``routine profile.'' A routine vehicle is one that does not have a wide variety of trips, while a non-routine vehicle has a diverse range of them. To illustrate, a cliché example of a routine vehicle would be one driven by someone who lives very simply and only uses their car to get food and essentials every Sunday at noon. A cliché example of a non-routine vehicle would be one driven by a pizza delivery person who has a changing schedule and is always making different kinds of trips at all times of the day and week. This being said, it is expected that a routine vehicle will have a small dispersion of local anomaly scores, while a non-routine one will have a large dispersion. To return to the illustration, the person who only uses their car to get food and essentials every Sunday at noon would then have a much less dispersed distribution of their scores than the pizza delivery person. This is why we refer to the local score vector as a ``routine profile.''

In the global scheme, the anomaly score of a trip, which we refer to as ``global,'' is calculated by applying an AD algorithm to all vehicles in the telematics dataset at once. The global anomaly score of a trip thus reflects its anomaly degree with respect to all other trips in the dataset. In this case, a $200$ kilometer trip made on Tuesday at $3$ a.m. is likely to be assigned a high anomaly score because this kind of trip, using common sense and looking at Figure~\ref{fig:telematics_histograms}, is hardly ever made by the general population of vehicles. Conversely, a ten kilometer trip made on Monday at rush hour will probably be assigned a low global anomaly score. Along these lines, a trip with a high global anomaly score is one that bears little resemblance to trips typically made by the general population of vehicles. The set of global anomaly scores for a vehicle's trips forms a vector of real numbers, or an empirical distribution, that we call the ``peculiarity profile.'' A somewhat peculiar vehicle is one that makes trips considered rather anomalous by the AD algorithm (i.e., it has high anomaly scores), while a somewhat non-peculiar vehicle makes trips considered rather ordinary (i.e., it has low anomaly scores).

Let us consider a dataset where $n$ equally weighted observations (or instances) are described by $p$ numeric variables. We denote by $\boldsymbol{X}~=~(x_{ij})_{n\times p}$ the data matrix, where $x_{ij}$ is the value of the $i^\text{th}$ observation for the $j^\text{th}$ variable. We also denote by $\boldsymbol{x}_i = (x_{i1}, \dots, x_{ip}) \in \mathbb{R}^p$ the $i^\text{th}$ observation (or row) of this matrix. Note that the observations $\boldsymbol{x}_1, \dots, \boldsymbol{x}_n$ define a cloud of $n$ points in $\mathbb{R}^p$. In the anomaly detection context, one is interested in determining an anomaly score, which is a real number that reflects the extent to which an observation is considered anomalous, for each observation. 

\subsection{Mahalanobis' method}

A simple way of obtaining an anomaly score for each point $\boldsymbol{x}_i$ is to compute its distance from the centroid of the point cloud $\overline{\boldsymbol{x}} = 1/n\sum_{i=1}^n \boldsymbol{x}_i$. Indeed, the distance from the center of the distribution seems a natural way of measuring the anomaly level of an observation: the further an observation is from the center, the more it should be considered anomalous. For this purpose, standard Euclidean distance is not a suitable distance metric if the variables are on different scales. The Mahalanobis distance (\cite{mahalanobis1936generalized}) is more advisable to use and is actually a fairly popular way of performing unsupervised anomaly detection. Mahalanobis' method computes the Mahalanobis distance between a point $\boldsymbol{x}$ and a point cloud with centroid $\boldsymbol{\overline{x}}$ and covariance matrix $\Sigma$, given by
 \begin{align*}
    d_M(\boldsymbol{x}) = \sqrt{(\boldsymbol{x} - \boldsymbol{\overline{x}})^\top \Sigma^{-1} (\boldsymbol{x} - \boldsymbol{\overline{x}})},
 \end{align*}
which can be directly used as an anomaly score.

\subsection{Local Outlier Factor}

The local outlier factor (LOF) algorithm (\cite{breunig2000lof}) is based upon the concept of local density, where ``local'' is defined by the $k$ nearest neighbors of a point. It is worth mentioning that $k$ is a hyperparameter and thus, its value must be carefully chosen by the user. A point has a low (respectively high) local density if its neighborhood has a low (respectively high) concentration of points. To determine whether a point is an anomaly or not, one compares its local density with those of its neighbors. If the local density of the point is higher (respectively lower) than those of its neighbors, it will be given a low (respectively high) anomaly score.

To understand exactly how this method works, let us first introduce the concept of $k$-distance, which is simply the distance between a point and its $k^\text{th}$ nearest neighbor. One typically uses the Euclidean distance metric on previously scaled data. The $k$ nearest neighbors of $\boldsymbol{x}$, whose set of indices is denoted with $\mathcal{N}_k(\boldsymbol{x})$, are the set of points that lie at a distance of $k\text{-distance}(\boldsymbol{x})$ or less from $\boldsymbol{x}$. Note that in case of a tie (i.e., if more than one point is at a distance of exactly $k\text{-distance}(\boldsymbol{x})$ from $\boldsymbol{x}$), the set $\mathcal{N}_k(\boldsymbol{x})$ actually contains more than $k$ indices. 

The \textit{reachability distance} (RD) of a point $\boldsymbol{x}$ from another point $\boldsymbol{x}_i$ is defined as
\begin{align*}
    \text{RD}_k(\boldsymbol{x} \text{ from } \boldsymbol{x}_i) = \max\left\{k\text{-distance}(\boldsymbol{x}_i), d(\boldsymbol{x}, \boldsymbol{x}_i)\right\},
\end{align*}
where $d(\boldsymbol{x}, \boldsymbol{x}_i)$ is the distance between point $\boldsymbol{x}$ and point $\boldsymbol{x}_i$. It may perhaps be observed that ``reachability distance'' is a slight misuse of language because it is actually not a distance metric in the mathematical term because it lacks the symmetry property. In other words, the RD of $\boldsymbol{x}$ from $\boldsymbol{x}_i$ is the actual distance between the two points, but is at least the $k$-distance of $\boldsymbol{x}_i$. Therefore, if the two points are ``sufficiently'' close, the actual distance is replaced by the $k$-distance of $\boldsymbol{x}_i$. \cite{breunig2000lof} state that using the RD instead of a real distance metric helps to achieve more stable results.

Using the RD, one can define the \textit{local reachability density} (LRD) of a point $\boldsymbol{x}$, given by
\begin{align*}
    \text{LRD}_k(\boldsymbol{x}) = \frac{1}{\ave_{i \in \mathcal{N}_k(\boldsymbol{x})} \left\{\text{RD}_k(\boldsymbol{x} \text{ from } \boldsymbol{x}_i)\right\}}.
\end{align*}
The LRD of a point is thus the inverse of the arithmetic average reachability distance from its neighbors. If a point is easily ``reachable'' by its neighbors (i.e., if the average reachability distance is low), it means that the neighbors will be relatively close to the point, and then, taking the inverse, the LRD will be relatively high. Conversely, if a point is not easily ``reachable'' by its neighbors, the LRD will be low. To sum it up, the LRD of a point measures the concentration of points in its neighboorhood.

Then, the LRD of a point $\boldsymbol{x}$ is compared to those of its neighbors, and its LOF anomaly score may be derived with 
\begin{align*}
    \text{LOF}_k(\boldsymbol{x}) = \frac{\ave_{i \in \mathcal{N}_k(\boldsymbol{x})} \left\{\text{LRD}_k(\boldsymbol{x}_i)\right\}}{\text{LRD}_k(\boldsymbol{x})}.
\end{align*}
The LOF score of a point is therefore the arithmetic average LRD of its neighbors divided by its own LRD. If point $\boldsymbol{x}$ has a low LRD relative to its neighbors, it means that its neighborhood is sparse compared to those of its neighbors, and that it is an outlier. In that case, its LOF will be greater than $1$. Conversely, if point $\boldsymbol{x}$ has a high LRD relative to its neighbors, it is considered an inlier, and consequently, its LOF will be lower than $1$.

\subsection{Isolation Forest}

The isolation forest (IF) algorithm, developed in \cite{liu2008isolation}, detects anomalies by means of an ensemble of binary trees, called \emph{isolation trees}, built upon the data matrix. What one calls a tree is in fact the representation of the recursive splitting (or partitioning) in two parts of the variable space $\mathbb{R}^p$. IF deviates from the anomaly detection mainstream, where normal observations are first modeled before anomalies can be identified as observations that do not conform to the modeled distribution. In contrast, IF directly detects anomalies using the concept of isolation. Indeed, anomalies are few in number and are different from other data points (which means they are found in sparse regions of the variable space), making them easier to isolate relative to normal data points. 

In essence, an isolation tree works by recursively partitioning the variable space $\mathbb{R}^p$ into two parts until all observations are isolated. At each step, the splitting is done by randomly selecting a variable and a split value for that variable. Assuming all observations are distinct, the partitioning process thus ends when all observations are in their own leaf corresponding to a hyperrectangular region of $\mathbb{R}^p$ resulting from the partitioning. From then on, one can count the number of splitting steps required to isolate each observation $i$, denoted $h(\boldsymbol{x}_i)$, which can be interpreted as the path length within the tree to reach the leaf where observation $i$ lies starting from the root node. The shorter the path length, the easier it is to isolate the observation, and the higher its degree of anomaly.

Because a single tree may yield unstable results, IF builds $M$ isolation trees in order to create an ensemble (or forest), yielding $M$ path lengths $h_1(\boldsymbol{x}_i), \dots, h_M(\boldsymbol{x}_i)$ for each observation $i$. One could compute the average path length for all $M$ trees $\overline{h}(\boldsymbol{x}_i)$ for each observation $i$ and use it directly as an anomaly score. However, the authors standardize the average path length to improve comparison and interpretation. For each observation $i$, they first divide $\overline{h}(\boldsymbol{x}_i)$ by the expected path length for any observation of the dataset, which depends only on the sample size $n$, given by
\begin{align}
    c(n) = 2H(n-1) - \frac{2(n-1)}{n},
    \label{eq:bst}
\end{align}
where $H(i)$ is the harmonic number, which can be estimated with $H(i) = \ln(i) + 0.5772156649$. Equation~\ref{eq:bst} originates from the field of binary search trees (BST). Indeed, isolation trees have a structure equivalent to BSTs, and thus the expected path length of an observation is the same as the expected length of an unsuccessful search in a BST. The ratio of $h(\boldsymbol{x}_i)$ over $c(n)$ is thereafter exponentiated in a way to obtain an anomaly score between $0$ and $1$. The anomaly score for observation $i$ is hence given by
\begin{align}
    s(\boldsymbol{x}_i, n) = 2^{-\frac{\overline{h}(\boldsymbol{x}_i)}{c(n)}}.
\end{align}
The higher the score $s$, the more the observation is considered anomalous.

It is worth noting that trees that make up an IF are usually not built on the entire dataset. IF is in fact known to work well when the sample size is small. Consequently, one usually chooses the number of instances $b$ to pick every time a tree is built.

\section{Supervised Binary Classification Model}\label{sec:Supervised Binary Classification Model}

Every supervised learning task requires $n$ labeled observations (or examples, or samples) gathered in a training dataset $\mathcal{D} = \{(\boldsymbol{x}_i, y_i)\}_{i=1}^n$, where $\boldsymbol{x}_i = (x_{i1}, \dots, x_{ip})\in \mathcal{X}$ is the $p$-dimensional input (or feature) vector for the $i^\text{th}$ observation and $y_i \in \mathcal{Y}$, the corresponding response (or label, target, output). The sets $\mathcal{X}$ and $\mathcal{Y}$ are often referred to as the feature space and the output space, respectively. The distinction between the features and the response results from the fact that the former are assumed to be relatively easy to collect, whereas the latter is harder or impossible to obtain in due course. For instance, in auto insurance, pricing features are typically policy-related attributes that are easily gathered through a form such as the the policyholder's gender, age and region, as well as the vehicle's age, model, brand, etc. Conversely, the response, which is often the number or cost of claims, is not available when one needs to apply the supervised learning algorithm: it is only known at the very end of the coverage period.

Supervised learning theory assumes that observations from dataset $\mathcal{D}$ are realizations of the random vector $(\boldsymbol{X}, Y)$, so that a joint probability distribution $p_{\boldsymbol{X}, Y}(\boldsymbol{x}, y) = p_{\boldsymbol{X}}(\boldsymbol{x}) p_{Y|\boldsymbol{X}}(y|\boldsymbol{x})$ exists. The goal for a new observation is to predict its response $y$ given its features $\boldsymbol{x}$ as accurately as possible. This is accomplished by training a supervised learning algorithm on the training dataset, which will ``learn'' a function that maps the features to the response as well as possible. Mathematically, a supervised learning algorithm seeks to find a member $h: \mathcal{X} \xrightarrow{} \mathcal{A}$ of a predefined hypothesis function space $\mathcal{H}$ such that $h(\boldsymbol{x})$ is as close as possible to $y$ for all observations, where $\mathcal{A}$ is the set of all possible predictions. In order to define ``closeness'' and thus properly choose $h$, a \emph{loss function} $\ell: \mathcal{A}\times \mathcal{Y} \xrightarrow{} \mathbb{R}^+$ is defined, where $\ell(y, h(\boldsymbol{x}))$ measures the distance between the prediction $h(\boldsymbol{x})$ and the actual response $y$. It is worth noting that this loss function must be adapted to the supervised learning problem being tackled. The goal is usually to find $h\in\mathcal{H}$ that minimizes the \emph{expected loss}, often referred to as the \emph{risk} $\mathcal{R}$ of the function $h$:
\begin{align*}
    \mathcal{R}(h) &= \Esp{\ell\left(Y, h(\boldsymbol{X})\right)}\\
    &= \int_{\mathcal{X} \times \mathcal{Y}} \ell\left(y, h(\boldsymbol{x})\right) p_{\boldsymbol{X}, Y}(\boldsymbol{x}, y)d\boldsymbol{x}dy.
\end{align*}
However, unless one has simulated the data, the joint distribution $p_{\boldsymbol{X}, Y}$ is unknown, making the risk $\mathcal{R}$ impossible to compute. Therefore, one typically minimizes the \emph{empirical risk} over the training dataset, given by
\begin{align*}
    \widehat{\mathcal{R}}(h) = \frac{1}{n} \sum_{i=1}^n \ell(y, h(\boldsymbol{x})).
\end{align*}
From then on, supervised learning algorithms use optimization algorithms to find a function $h$ that minimizes, globally or locally, the empirical risk function. We find ourselves specifically in the context of \emph{supervised binary classification} when the response can take two distinct values (often encoded with $0$ and $1$), namely when $\mathcal{Y} = \{0, 1\}$. 

\subsection{Logistic Regression}\label{subsection:lr}

Logistic regression is among the most popular algorithms for supervised binary classification problems. It assumes that the conditional distribution of the response given the features is a Bernoulli distribution, which is
\begin{align*}
    p_{Y_i|\boldsymbol{X}_i}(y_i|\boldsymbol{x}_i) = \pi_i^{y_i} (1 - \pi_i)^{1 - y_i}, \quad y_i = 0, 1,
\end{align*}
where $\pi_i \defeq \mathbb{P}(Y_i = 1|\boldsymbol{X}_i = \boldsymbol{x}_i)$ is the conditional probability of being in class ``1'' for observation $i$. It further assumes that $\pi_i$ can be expressed as the sigmoid transform $\sigma(\cdot)$ of the linear predictor $\boldsymbol{x}_i\boldsymbol{\beta}$, where $\boldsymbol{\beta} = (\beta_0, \dots, \beta_p)$ is a parameter vector:
\begin{align*}
    \pi_i = \frac{1}{1 + \exp{(-\boldsymbol{x}_i \boldsymbol{\beta}})} \defeq \sigma(\boldsymbol{x}_i\boldsymbol{\beta}).
\end{align*}
Notice that the dot product $\boldsymbol{x}_i\boldsymbol{\beta}$ seems undefined here because $\boldsymbol{x}_i \in \mathbb{R}^p$ and $\boldsymbol{\beta} \in \mathbb{R}^{p+1}$. However, in order to allow for an intercept $\beta_0$ in the model, the feature vector is most often extended with the value ``$1$'', so that $\boldsymbol{x}_i = (1, x_{i1}, \dots, x_{ip})\in \mathbb{R}^{p+1}.$ Therefore, logistic regression seeks to find a function, which takes the form of the sigmoid transform of a linear transformation of the predictors, mapping the features $\boldsymbol{x}$ to a probability. The hypothesis space is thus $\mathcal{H} = \{\boldsymbol{x} \mapsto \sigma(\boldsymbol{x}\boldsymbol{\beta})|\boldsymbol{\beta}\in \mathbb{R}^p\}$, and the set of possible predictions, $\mathcal{A} = [0, 1]$. From then on, a function $h\in\mathcal{H}$ must be chosen, which is equivalent to choosing (or estimating) the parameters $\boldsymbol{\beta}$. This is often done by minimizing the empirical risk with binary cross-entropy loss, given by
\begin{align}
    \widehat{\mathcal{R}}(\boldsymbol{\beta}) = -\frac{1}{n} \sum_{i=1}^n y_i\ln(\pi_i) + (1 - y_i)\ln(1 - \pi_i).
    \label{eq:cross_entropy_risk}
\end{align}
It is noteworthy that minimizing the empirical risk in Equation~\ref{eq:cross_entropy_risk} is equivalent to estimating the parameters by maximum likelihood, because the right-hand side of the equation is just the Bernoulli log-likelihood times minus one. The estimated parameter vector that minimizes (\ref{eq:cross_entropy_risk}), denoted as $\widehat{\boldsymbol{\beta}}^\text{MLE}$, does not have a closed-form solution, but can be computed with a variety of numerical optimization methods, such as the method of iteratively reweighted least squares. Once estimated, the parameters can be used to obtain a prediction for a new observation $i= 0$:
\begin{align}
    \widehat{\pi}_0 = \sigma\left(\boldsymbol{x}_0 \widehat{\boldsymbol{\beta}}^\text{MLE}\right).
\end{align}

\subsection{Elastic-Net Regularization}\label{sub:elasticnet}

Having low bias and low variance predictions are two desirable properties of a supervised learning model. Indeed, one can break down the reducible error of a prediction model as the sum of bias and variance. From the well-known \emph{bias-variance tradeoff} in machine learning, we know that one can decrease the bias of a model by increasing its variance, and vice-versa. It can be shown that the parameter vector's maximum likelihood estimator $\widehat{\boldsymbol{\beta}}^\text{MLE}$ of Subsection~\ref{subsection:lr} is the asymptotically unbiased estimator of the true parameter vector $\boldsymbol{\beta}$ that has the smallest variance. However, it rarely is the best estimator for prediction because it still exhibits large variance, as do the resulting predictions. Regularization adds a regularization (or penalty) term to the empirical risk, which reduces variance at the cost of increasing bias. In many cases, the decrease in variance more than offsets the increase in bias, which improves prediction performance. Various types of regularization exist, including the well-known Ridge and Lasso regressions. 

In the case of Ridge-regularized logistic regression, a penalty term that is proportional to the sum of squared coefficients is added to the empirical risk. The optimization problem hence becomes
\begin{align}
    \boldsymbol{\widehat{\beta}}^\text{RIDGE} = \argmin_{\boldsymbol{\beta}} \left\{ \widehat{\mathcal{R}}(\boldsymbol{\beta}) + \lambda\sum_{j=1}^p \beta_j^2 \right\},
    \label{eq:ridge}
\end{align}
where $\lambda\ge 0$ is a hyperparameter that controls the amount of regularization. Examining Equation~\ref{eq:ridge}, one notices that the algorithm must not only minimize the empirical risk, but also the magnitude of the parameters, which has the effect of shrinking them toward zero. The resulting estimates (and thus the predictions) are biased because they are being pushed toward zero from the unbiased maximum likelihood estimates. In return, predictions have a lower variance because they result from smaller parameters (in absolute value). One should note that the intercept parameter $\beta_0$ is not included in the penalty term in Equation~\ref{eq:ridge}: this is because no shrinkage of its estimated coefficient is desired because it does not reflect the impact of a predictor on the response. However, the intercept parameter is most often omitted because it is customary and recommended to center (and scale) the data prior to modeling. The larger the $\lambda$ hyperparameter, the more biased the estimators will be and the less their variance will be. The aim is to find a good tradeoff between bias and variance by tuning the value of $\lambda$. Its value is not directly optimized by the algorithm, so one must choose it carefully based on out-of-sample performance, like with cross-validation. For a given value of $\lambda$, the optimization problem of Equation~\ref{eq:ridge} is convex and can therefore easily be solved using numerical optimization methods.

Although Ridge regression shrinks the coefficients towards zero, it never actually sets them to zero, yielding non-parsimonious models. Put another way, Ridge regression has no built-in feature selection mechanism, which is a flaw because sparser models are more easily interpreted by humans. Lasso regression, on the other hand, has the ability to set coefficients to zero due to the nature of its regularization term, and thus allows for automatic feature selection. Lasso works in a similar way as the Ridge regression except that the penalty term is proportional to the sum of the \emph{absolute values} of the coefficients, and the optimization problem thus becomes
\begin{align}
    \boldsymbol{\widehat{\beta}}^\text{LASSO} = \argmin_{\boldsymbol{\beta}} \left\{ \widehat{\mathcal{R}}(\boldsymbol{\beta}) + \lambda\sum_{j=1}^p |\beta_j| \right\}.
    \label{eq:lasso}
\end{align}
As with Ridge regression, one wants to find a good tradeoff between bias and variance by tuning the $\lambda$ hyperparameter. Once the value is set for $\lambda$, the convex optimization problem in (\ref{eq:lasso}) can be solved fairly easily using convex optimization theory. 

\begin{figure}
    \centering
    \includegraphics[width = \textwidth]{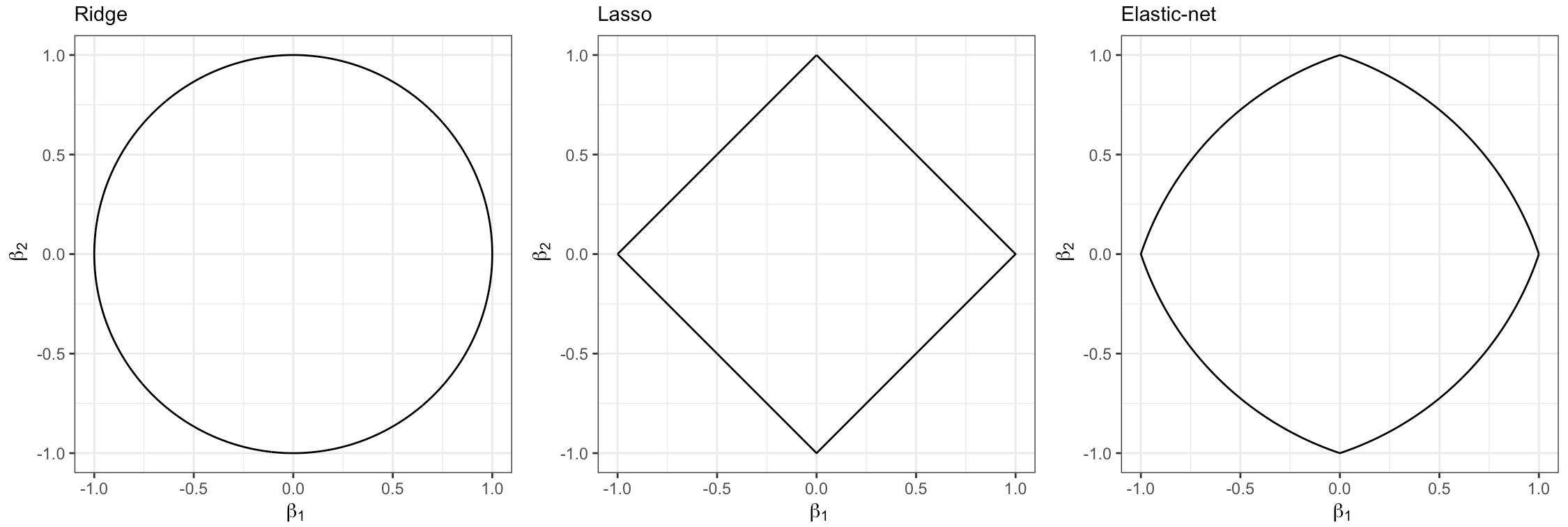}
    \caption{Boundary of the constraint regions in two dimensions (i.e. $\boldsymbol{\beta} = (\beta_1, \beta_2)$) for Ridge, lasso and elastic-net regressions, for $s = 1$. For the elastic-net regression, we set $\alpha = 0.5$.}
    \label{fig:constraint_regions}
\end{figure}

Equations~(\ref{eq:ridge})~and~(\ref{eq:lasso}) are called the Lagragian formulations of Ridge and lasso logistic regressions, respectively. In order to better understand the impact of both types of penalties, it is useful to represent them both with a constrained optimization problem, given by
\begin{align}
    \boldsymbol{\widehat{\beta}} = \argmin_{\boldsymbol{\beta}} \left\{ \widehat{\mathcal{R}}(\boldsymbol{\beta})\right\} \quad \text{s.t}\quad \sum_{j=1}^p |\beta_j|^q \le s,
    \label{eq:constrained}
\end{align}
where $q=1$ and $q=2$ correspond respectively to the lasso and Ridge cases, and where $s$ is a hyperparameter with a one-to-one correspondence to the $\lambda$ hyperparameter of the Lagrangian formulation. It is easier to remark with this constrained formulation that regularization gives a coefficient ``budget'' to the model. Indeed, the aim is to minimize the empirical risk while staying within the budget of $s$ for the $\ell_q$-norm (raised to the power of $q$) $||\boldsymbol{\beta}||_{\ell_q}^q = \sum_{j=1}^p |\beta_j|^q$ of the parameter vector. The constraint in Equation~\ref{eq:constrained} actually defines a hypersolid (or constraint region) in $\mathbb{R}^p$ in which the estimated parameter vector must lie. In the special case of Ridge regression (i.e. $q = 2$), the constraint region is actually a hyperball, while for lasso (i.e. $q = 1$), it is a polytope. In the logistic regression framework with cross-entropy as a loss function, the empirical risk is convex and thus has the shape of a $p$-dimensional infinite bowl. Consequently, the solution to the optimization problem will always lie on the boundary of the constraint region, which means one looks for the point $\widehat{\boldsymbol{\beta}}$ where the bowl-shaped empirical risk function intersects with the constraint region. With this in mind, the shape of the hypersolid can help in understanding the behavior of regularized models. Indeed, the rationale for Ridge not performing automatic feature selection is that its constraint region has no sharp corners, being a hyperball. Therefore, the intersection point will almost surely not touch one of the axes which means that no coefficient will be set to zero. In contrast, lasso performs feature selection because a polytope has sharp corners. In two dimensions, namely if $\boldsymbol{\beta} = (\beta_1, \beta_2)\in\mathbb{R}^2$, the border of the constraint region is a circle in the Ridge case and a rhombus in the lasso case, and they are illustrated in Figure~\ref{fig:constraint_regions}. 

Lasso regression, however, is known to handle groups of features that are highly correlated with one another poorly due to the sharp corners of its constraint region. Indeed, for a group of strongly correlated features, lasso tends to choose only one of them in the group and does not care which one is selected. In contrast, Ridge will select all features and share the coefficient ``budget'' approximately equally among them, which is more desirable. Elastic-net regularization (\cite{zou2003regression}) compromises between Ridge and lasso penalties, and solves the following optimization problem:
\begin{align}
    \boldsymbol{\widehat{\beta}}^\text{E-N} = \argmin_{\boldsymbol{\beta}} \left\{ \widehat{\mathcal{R}}(\boldsymbol{\beta}) + \lambda\left[\left(1-\alpha\right)\sum_{j=1}^p \beta_j^2 + \alpha \sum_{j=1}^p |\beta_j|\right] \right\},
    \label{eq:en}
\end{align}
where $\alpha \in [0, 1]$ is a ``mixing'' hyperparameter that controls the tradeoff between Ridge and lasso penalties. Alternatively, one can solve the following constrained optimization problem:
\begin{align}
    \boldsymbol{\widehat{\beta}}^\text{E-N} = \argmin_{\boldsymbol{\beta}} \left\{ \widehat{\mathcal{R}}(\boldsymbol{\beta})\right\} \quad \text{s.t} \quad \left(1-\alpha\right)\sum_{j=1}^p \beta_j^2 + \alpha \sum_{j=1}^p |\beta_j| \le s, 
    \label{eq:en_constrained}
\end{align}
where $s$ is a hyperparameter analogous to the $\lambda$ hyperparameter of Equation~\ref{eq:en}. In Figure~\ref{fig:constraint_regions}, one notices that the elastic-net constraint region shares both Ridge and lasso properties, having both rounded edges and sharp corners. The former allows for feature selection, while the latter helps to cope with highly correlated features by sharing the coefficient budget among them. Elastic-net regularization thus combines the best of both worlds. Moreover, it solves a convex program, which means the parameters can be estimated quite readily using convex optimization techniques. 

In addition to yielding parsimonious and interpretable models as well as handling collinearity effectively, elastic-net regression has another major benefit. Indeed, elastic-net regression can be considered an ``off-the-shelf'' algorithm, meaning that it can be applied to data and quickly achieve good results. In point of fact, elastic-net regression requires very little data preprocessing and has very few hyperparameters (only two: $\alpha$ and $\lambda$), meaning they are easily tunable. In contrast, some neural networks and boosting algorithms can include dozens of hyperparameters, all of which must be carefully tuned. For further information about elastic-net regularization (or more generally, regularization), we refer to \cite{hastie2015statistical}.

\subsection{Preprocessing}

\subsubsection{Anomaly scores}\label{subsection:anomalyscores}

First of all, let us recall that each of the anomaly detection algorithms presented in Section~\ref{sec:Anomaly Detection Algorithms} are used to compute both a local and a global anomaly score, which is a real number, for each vehicle trip in the telematics dataset. Once these scores are computed, which constitute the vehicles' routine and peculiarity profiles, the goal is to use them in conjunction with traditional risk factors and distance driven in an elastic-net logistic regression model to perform claim classification. Indeed, we expect routine and peculiarity profiles to convey important information to explain the occurrence of claims. For a given anomaly detection technique, each vehicle is associated with as many anomaly scores as the number of trips made by it. Obviously, not all vehicles completed the same number of trips during their one-year observation period, meaning their routine or peculiarity profile vector has a variable length. Consequently, anomaly scores cannot be entered directly as covariates in the elastic-net model, for the latter only accepts rectangular design matrices as inputs. It is therefore needed to extract features from the routine and peculiarity profiles of the vehicles. For this purpose, in a way that preserves information as much as possible for each vehicle, we extract $11$ quantiles of its anomaly score vector that are evenly spread out, namely the $0^\text{th}, 10^\text{th}, \dots, 90^\text{th}$ and $100^\text{th}$ percentiles. The $0^\text{th}$ and $100^\text{th}$ percentiles are also referred to as the minimum and the maximum of the anomaly score vector, respectively. This way, each vehicle's telematics data is summarized with $11$ real numbers, meaning each vehicle ends up with $11$ extracted telematics features, which will later be used to enhance claim classification. The preprocessing of the anomaly scores is illustrated in Table~\ref{tab:preproc_ano_scores}.
\begin{table}[ht]
    \centering
    \begin{adjustbox}{max width = \textwidth}
        \begin{tabular}{c c c}
            \toprule 
            \textbf{VIN} & \textbf{Trip ID} & \textbf{Anomaly score}\\ 
            \midrule
            A & $1$ & $\mathbb{R}$\\
            A & $2$ & $\mathbb{R}$\\
            $\vdots$ & $\vdots$ & $\vdots$\\
            A & $2320$ & $\mathbb{R}$\\
            \midrule
            B & $1$ & $\mathbb{R}$\\
            B & $2$ & $\mathbb{R}$\\
            $\vdots$ & $\vdots$ & $\vdots$\\
            B & $1485$ & $\mathbb{R}$\\
            \midrule
            C & $1$ & $\mathbb{R}$\\
            $\vdots$ & $\vdots$ & $\vdots$\\
            \bottomrule 
        \end{tabular}
        
        $\xRightarrow[\text{percentiles}]{\text{extract}}$
        
        \begin{tabular}{c c c c c c}
            \toprule
            & \multicolumn{5}{c}{\textbf{Anomaly score's percentiles}}\\
            \cmidrule(l){2-6}
            \textbf{VIN} & $0^\text{th}$ & $10^\text{th}$ & \dots & $90^\text{th}$ & $100^\text{th}$\\ 
            \midrule
            A & $\mathbb{R}$ & $\mathbb{R}$ & \dots & $\mathbb{R}$ & $\mathbb{R}$\\
            B & $\mathbb{R}$ & $\mathbb{R}$ & \dots & $\mathbb{R}$ & $\mathbb{R}$\\
            C & $\mathbb{R}$ & $\mathbb{R}$ & \dots & $\mathbb{R}$ & $\mathbb{R}$\\
            \bottomrule 
        \end{tabular}
    \end{adjustbox}
    \caption{Example of anomaly score preprocessing from the extract of Table~\ref{tab:1}.} 
    \label{tab:preproc_ano_scores}
\end{table}

\subsubsection{Features}\label{sub:features}

Although elastic-net logistic regression can be qualified as an off-the-shelf model, the features still need a few preprocessing steps before they can be entered into the classification algorithm, either for training or scoring. Features include the ten traditional risk factors from Table~\ref{tab:classic}, the distance driven and the $11$ telematics features extracted in Subsection~\ref{subsection:anomalyscores}, which makes for a total of $22$ features. Each of them undergoes different preprocessing steps according to their nature. To carry out this preprocessing efficiently, the very handy \texttt{recipes} package from the \texttt{R} programming language is used. This package allows the definition of a preprocessing ``recipe'' (or workflow) consisting of multiple preprocessing steps to apply every time a model is either trained or scored. The \texttt{recipes} package can be used to center/scale features, impute missing values, numerically encode and group modalities for categorical features, etc. A key benefit of using this package is that a preprocessing recipe can be easily paired with a predictive model, thus avoiding data leakage. Indeed, it is not advisable to preprocess the whole dataset once at the beginning of the modeling pipeline because in such a case, information from the testing set may leak into the training set, which could lead to an overestimation of the model's performance. Similarly, information from the validation fold could leak into the training folds while cross-validating. In order to properly assess generalization power, the testing/validation set must indeed be completely disjoint from the training set. This is why it is best to make preprocessing an integral part of the modeling process by treating preprocessing and the prediction model as a single entity. That way, the preprocessing steps are performed every time a model is either trained or scored, and not using data that should be kept for testing/validation. 

\paragraph{Categorical features}

Just like many other supervised learning algorithms, elastic-net logistic regression is unable to deal with categorical features as is; they must first be numerically encoded. The categorical features used in the classification model include the four numerical traditional risk factors from Table~\ref{tab:classic}, namely \texttt{gender}, \texttt{marital\_status}, \texttt{pmt\_plan} and \texttt{veh\_use}. For this purpose, target encoding is employed, which uses the target (or response) variable to assign a real number to each of the feature's categories. This is done by fitting a non-penalized logistic regression model without an intercept term on the response variable (which is the indicator of a claim in the observation period) using the feature to be encoded as the only covariate. This results in a coefficient for every category that is indicative of the claim risk for that category. These coefficients are then directly used as encoding values. This type of encoding is an alternative to the more common binary (or dummy) encoding, which creates several dummy variables from a categorical feature that indicates whether or not the observation belongs to each category. Conversely, target encoding has the benefit of not increasing the dimensionality of the dataset. Note that prior to target encoding, rare categories, namely those whose occurrence in the data is 5\% of the observations or less, are pooled in an ``other'' category. Once they have been converted to numeric data, categorical features still need a few preprocessing steps related to numerical features, described in the next paragraph.
\begin{Example}[Target encoding]
    Let us consider the small fictitious dataset in Table~\ref{tab:target_encoding}, with only one feature $x$ and a binary response $y$. Suppose one wishes to target encode feature $x$, which is categorical with three categories: ``blue'', ``white'' and ``red''. Using non-penalized logistic regression without an intercept term, one first obtains the three coefficients by minimizing the empirical risk in Equation~\ref{eq:cross_entropy_risk} or, equivalently, by maximizing likelihood: $\widehat{\beta}_{blue} = 0$, $\widehat{\beta}_{white} = -20.57$, $\widehat{\beta}_{red} = 20.57$. These estimated coefficients are then used to encode $x$.
    \begin{table}[ht]
        \centering
        \begin{tabular}{c | c c}
            \toprule
            $y$ & $x$ & $x_{encoded}$\\
            \midrule
            $1$ & red & $20.57$\\
            $0$ & blue & $0$\\
            $1$ & red & $20.57$\\
            $1$ & blue & $0$\\
            $0$ & white & $-20.57$\\
            \bottomrule 
        \end{tabular}
        \caption{Example of target encoding with a non-penalized logistic regression on a fictitious dataset.}
        \label{tab:target_encoding}
    \end{table}
\end{Example}

\paragraph{Numerical features}

\begin{figure}
    \centering
    \includegraphics[width = \textwidth]{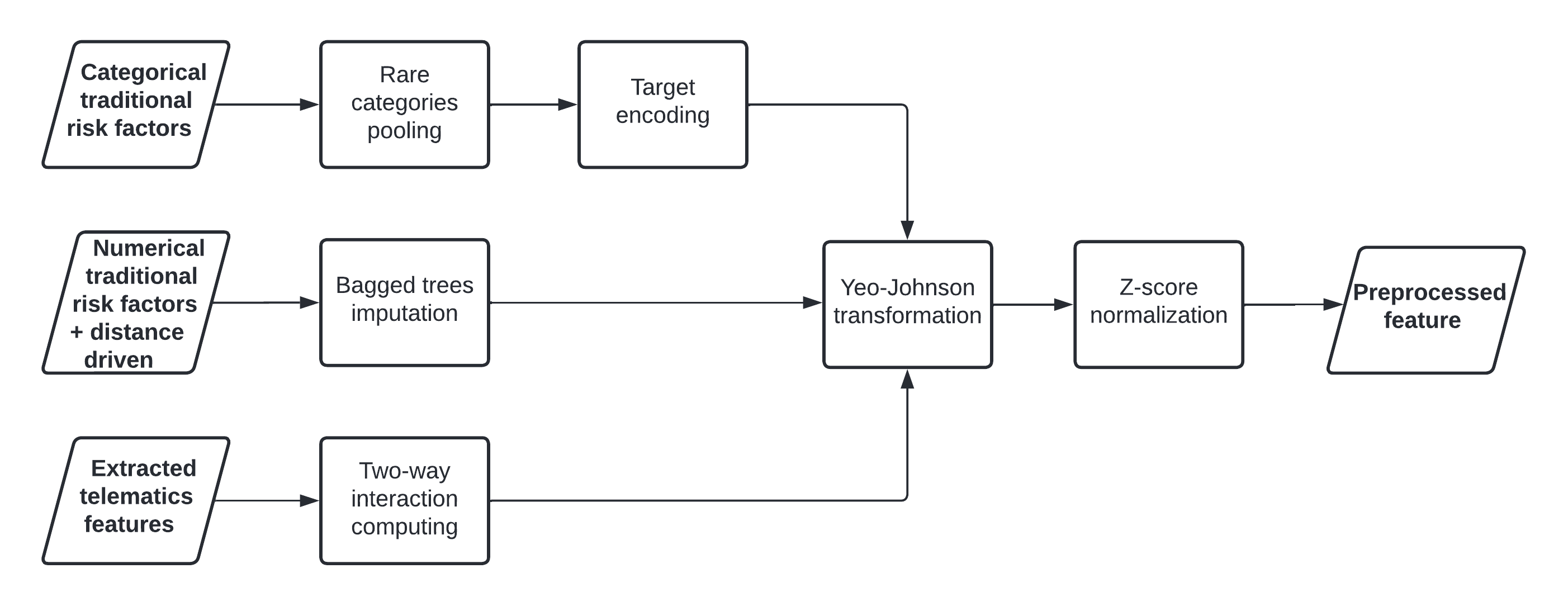}
    \caption{Flowchart of the feature preprocessing workflow, which is applied every time a model is either trained or scored.}
    \label{fig:preprocessing_recipe}
\end{figure}

The numerical features used in the classification model include six of the features in Table~\ref{tab:classic} (\texttt{annual\_distance}, \texttt{commute\_distance}, \texttt{cov\_count\_3\_yrs\_minor}, \texttt{veh\_age}, \texttt{years\_claim\_free} and \texttt{years\_licensed}), distance driven, the four numerically encoded categorical features of Table~\ref{tab:classic} (\texttt{gender}, \texttt{marital\_status}, \texttt{pmt\_plan} and \texttt{veh\_use}) and the $11$ quantiles extracted from the routine and peculiarity profiles in Subsection~\ref{subsection:anomalyscores}. To better capture the vehicles' routine and peculiarity profiles, all $\binom{11}{2} = 55$ degree two interactions between the $11$ anomaly scores-based features are first computed. All $6 + 4 + 11 + 55 = 76$ numerical features then undergo a Yeo-Johnson transformation followed by a z-score normalization. The Yeo-Johnson transformation reduces the skewness of the features' distributions and gets them closer to a normal distribution, which is usually beneficial for supervised learning algorithms. Mathematically, the Yeo-Johnson transformation $\psi$ of $x\in\mathbb{R}$ is defined as
\begin{align}
    \psi(x, \lambda) =
    \begin{cases}
      ((x + 1)^\lambda - 1)/\lambda & \text{if } \lambda \ne 0, x \ge 0\\
      \ln(x + 1) & \text{if } \lambda = 0, x \ge 0\\
      -[(-x + 1)^{2 - \lambda} - 1]/(2 - \lambda) & \text{if } \lambda \ne 2, x < 0\\
      -\ln(-x + 1) & \text{if } \lambda = 2, x < 0, \\
    \end{cases}
\end{align}
where $\lambda$ is a parameter that is usually optimized by maximum likelihood such that the empirical distribution of the transformed values is as close as possible to the normal distribution. Z-score normalization has the effect of centering and scaling the feature vectors. ltures allows us to omit the intercept parameter $\beta_0$ in the elastic-net model, while scaling them ensures they all have equal importance in the modeling process. It is strongly recommended to scale features prior to feeding them into regularized logistic regression models because they are not scale invariant. Consider the feature vector $\boldsymbol{x} = (x_1, \dots, x_n)$ with empirical mean $\overline{x} = \frac{1}{n}\sum_{i=1}^n x_i$ and standard deviation $s = \frac{1}{n-1}\sum_{i=1}^n (x_i - \overline{x})^2$. The z-score normalized version of $\boldsymbol{x}$ is
\begin{align}
    \boldsymbol{x}^* = \left(\frac{x_1 - \overline{x}}{s}, \dots, \frac{x_n - \overline{x}}{s}\right).
\end{align}
Note that all features are complete except \texttt{commute\_distance} from Table~\ref{tab:classic}, which has a missing rate of $21.5$\%. The missing values are imputed using bagged trees imputation, which first trains a committee of $25$ regression trees on all instances having a non-missing value for  \texttt{commute\_distance} feature. It then substitutes missing values with the predictions produced by the tree committee. The complete data preprocessing workflow is illustrated in Figure~\ref{fig:preprocessing_recipe}.

\section{Hyperparameter tuning}\label{sec:Hyperparameter tuning}

\subsection{Anomaly Detection Techniques}\label{ssec:Anomaly Detection Techniques}

Among the three AD algorithms presented in Section~\ref{sec:Anomaly Detection Algorithms}, LOF and IF have hyperparameters that need to be tuned. Indeed, with the LOF algorithm, one must choose the size of the neighborhood, defined by $k$, while in the case of IF, one must choose a value for the sampling size $b$. These hyperparameters are not directly optimized by their respective algorithm, so their value must be carefully chosen in accordance with the goal to achieve in mind. Our goal is to obtain the best possible claim classification performance and to this end, hyperparameter values are chosen in such a way to optimize the area under the receiver operating characteristic curve (AUC), a popular threshold-free binary classification performance metric, when anomaly scores-based features are used alone in a non-penalized logistic regression model. Recall that anomaly scores are computed in two different ways, namely the local way, where anomaly detection is applied separately on each vehicle and the global way, where the latter is applied once on the whole portfolio. Therefore, four hyperparameters need optimization, namely $k$ and $b$ for both local and global schemes. Hyperparameter tuning is done using grid search, and AUC is obtained using five-fold cross-validation.

\subsubsection{Local scheme}

In the local scheme, the neighborhood size in the LOF algorithm and the sampling size in the IF algorithm are expressed as a fraction of the number of trips for each vehicle. Therefore, $k_{frac}$ and $b_{frac}$, which represent respectively the fraction of trips used as the nearest neighbors in the LOF algorithm and the fraction of trips to draw every time an isolation tree is built in the IF algorithm, are tuned instead of $k$ and $b$. The relation between $k$ and $k_{frac}$ as well as between $b$ and $b_{frac}$ are given by
\begin{align}
    k_i &=  k_{frac} \times n_i,\\
    b_i &=  b_{frac} \times n_i,
\end{align}
where $n_i$ is the number of trips made by the $i^\text{th}$ vehicle. That way, algorithms are applied to each vehicle $i$ using its corresponding value $k_i$ (or $b_i$), which is proportional to its number of trips. The values tested for $k_{frac}$ and $b_{frac}$ go from $0.05$ to $0.6$ and from $0.05$ to $1$, respectively, in leaps of $0.05$. Therefore, the grids used for the grid search are $\mathcal{G}_{k_{frac}} = \{0.05i\}_{i=1}^{12}$ and $\mathcal{G}_{b_{frac}} = \{0.05i\}_{i=1}^{20}$. 

\begin{figure}
  \centering
  \subcaptionbox{Local LOF\label{fig:tune_local_lof}}
    {\includegraphics[width = \textwidth]{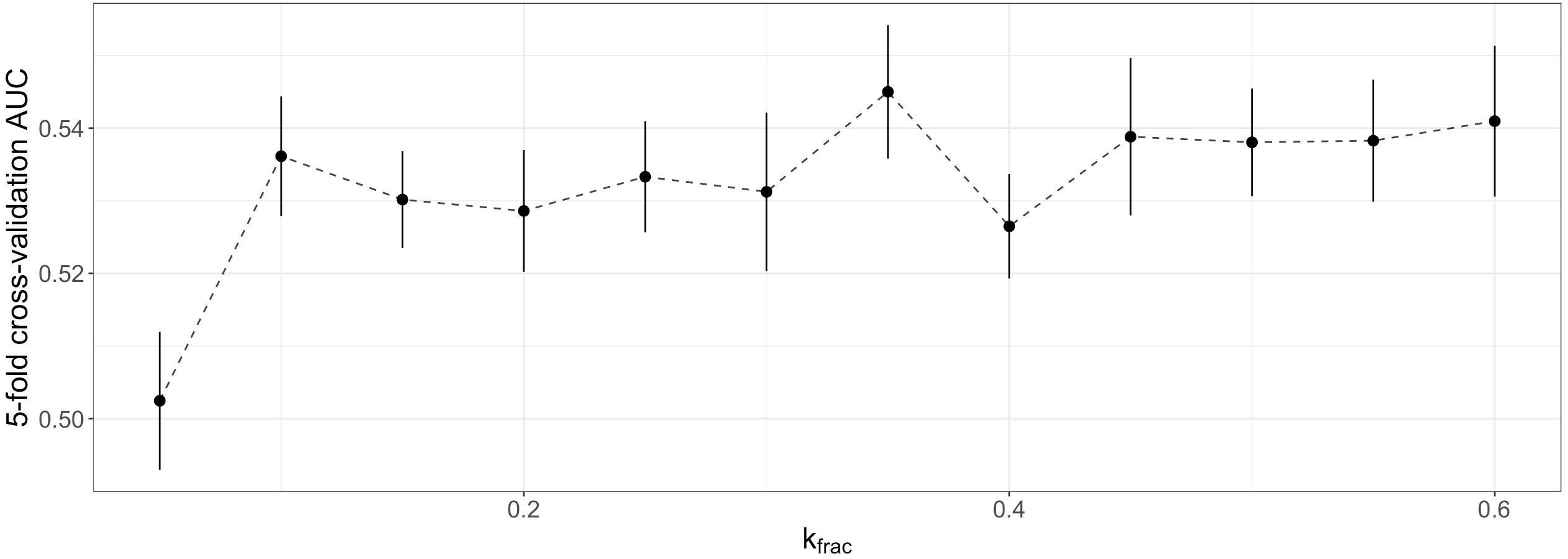}}
  \subcaptionbox{Local IF\label{fig:tune_local_if}}
    {\includegraphics[width = \textwidth]{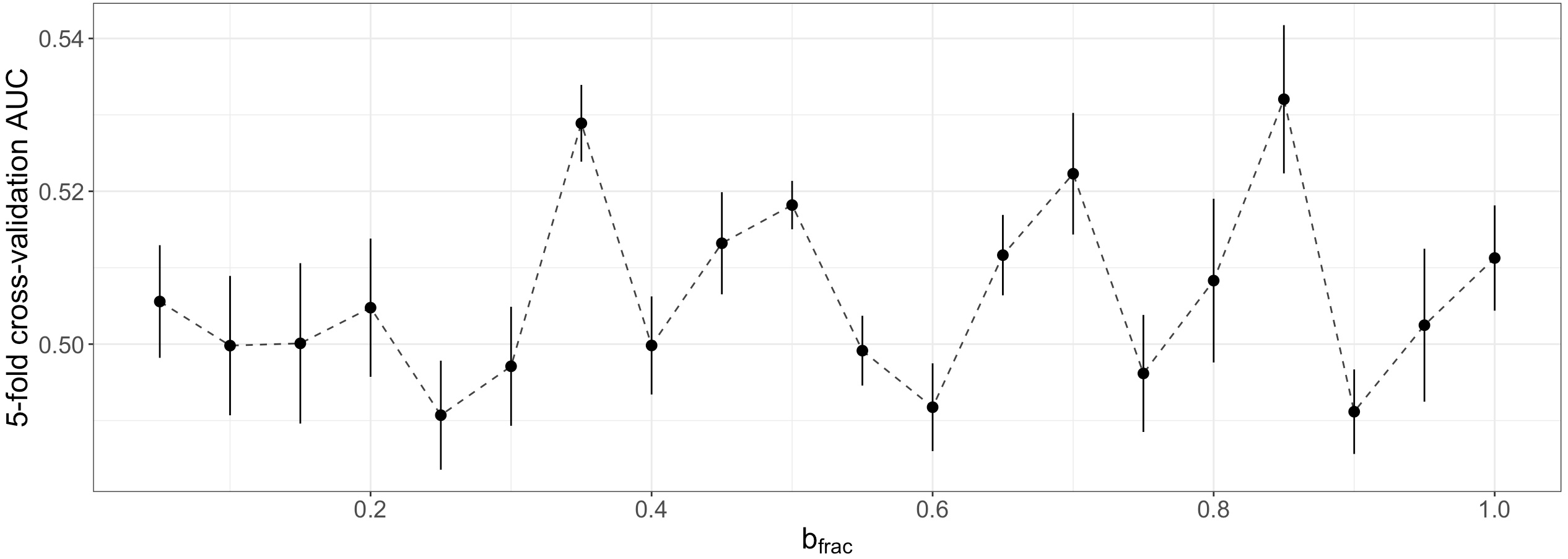}}
  \caption{Tuning results for local anomaly detection algorithms. The five-fold cross-validation AUC and its corresponding standard deviation (represented as the length of the vertical segment) is shown for each hyperparameter value.}\label{fig:tune_local_methods}
\end{figure}

An AUC value is obtained for each value in $\mathcal{G}_{k_{frac}}$ and $\mathcal{G}_{b_{frac}}$, whereupon the two chosen values (one for $k_{frac}$ and one for $b_{frac}$) are those that maximize AUC. For a given hyperparameter value in $\mathcal{G}_{k_{frac}}$ or $\mathcal{G}_{b_{frac}}$, the cross-validation AUC is obtained in the following way. First, the anomaly detection technique (either LOF or IF) is applied to each vehicle separately using the given hyperparameter value, yielding a local anomaly score for each trip and hence a routine profile for each vehicle. The $11$ telematics features are then extracted from the routine profile vector according to Subsection~\ref{subsection:anomalyscores}. Finally, after computing all $55$ two-way interactions between the $11$ telematics features, a five-fold cross-validation is performed using a non-penalized logistic regression, yielding an out-of-sample estimated probability of claiming for each vehicle in the training dataset. The AUC value is then computed by comparing the estimated probability vector with the response vector. The tuning results for the local scheme are shown in Figure~\ref{fig:tune_local_methods}.
It turns out that the optimal hyperparameter values that are found are $k_{frac} = 0.35$ for LOF and $b_{frac} = 0.85$ for IF, with respective AUCs of $0.5450$ and $0.5320$. These are the hyperparameter values that will be used from now on in the local scheme. It is worth noting that Mahalanobis' method, which has no hyperparameter, yields an AUC of $0.5200$. The optimal hyperparameter for each local anomaly detection method and its corresponding AUC value are given in Table~\ref{tab:tuning_anomaly}.

\subsubsection{Global scheme}

\begin{figure}
  \centering
  \subcaptionbox{Global LOF\label{fig:tune_global_lof}}
    {\includegraphics[width = \textwidth]{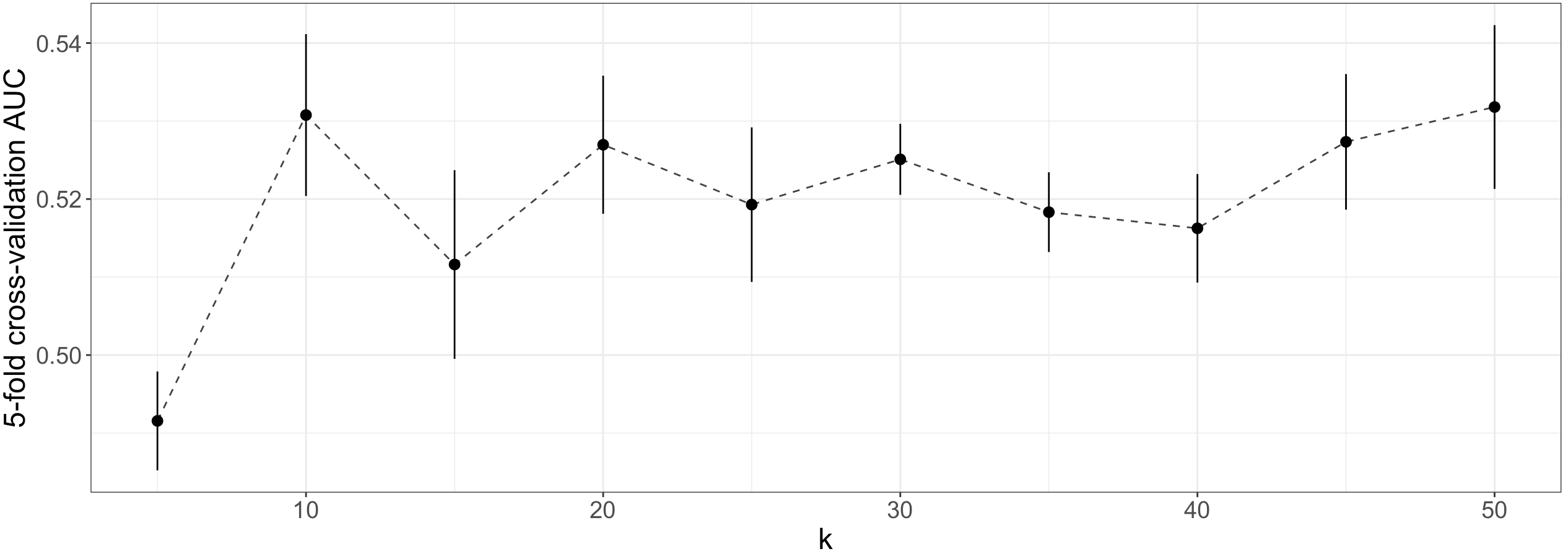}}
  \subcaptionbox{Global IF\label{fig:tune_global_if}}
    {\includegraphics[width = \textwidth]{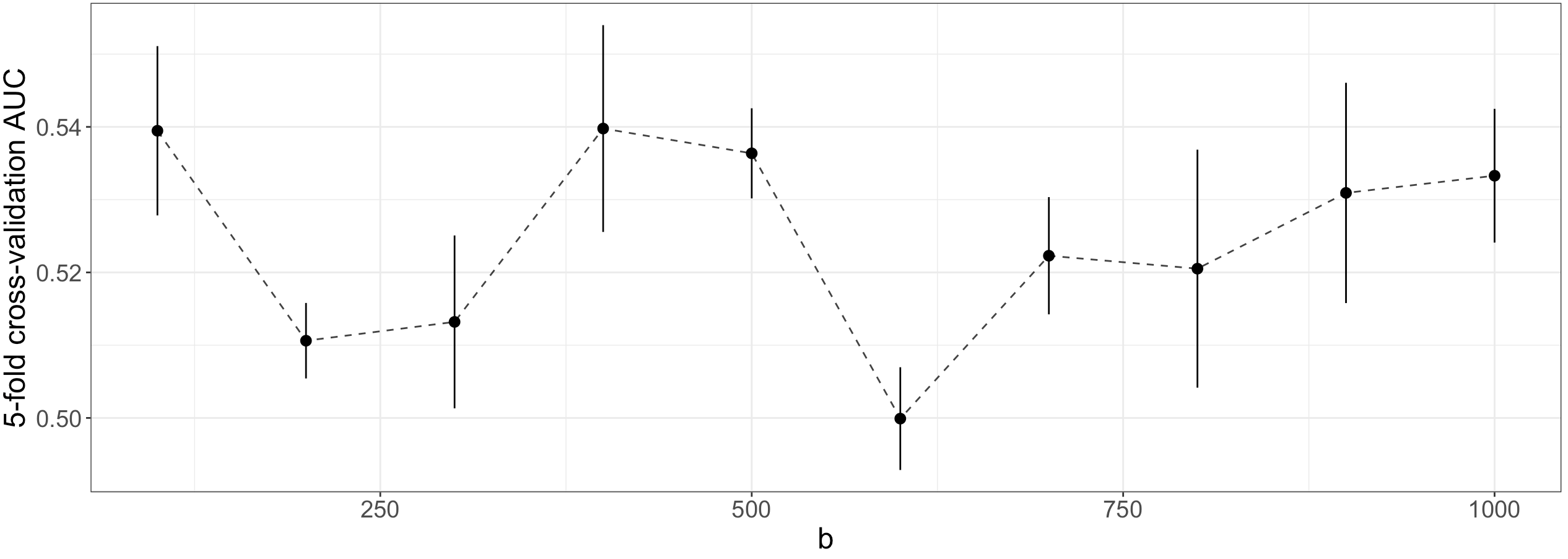}}
  \caption{Tuning results for global anomaly detection algorithms. The five-fold cross-validation AUC and its corresponding standard deviation (represented as the length of the vertical segment) is shown for each hyperparameter value.}\label{fig:tune_global_methods}
\end{figure}

In the global scheme, hyperparameter tuning is done in a similar way as in the local scheme with the exception that the neighborhood size and the sampling size are not expressed as a fraction of the number of trips, but as an actual number of trips. The hyperparmeters $k$ and $b$ are thus tuned directly. Again, a grid search is used with $\mathcal{G}_k = \{5i\}_{i=1}^{10}$ and $\mathcal{G}_b =  \{100i\}_{i=1}^{10}$ as tuning grids, and the chosen value for $k$ (and for $b$) is the one matching the highest cross-validation AUC. In order to obtain an AUC for a specific value in $\mathcal{G}_k$ or $\mathcal{G}_b$, the anomaly detection technique (either LOF or IF) is applied on the whole training dataset using this specific value, resulting in a global anomaly score for each trip. In the same way as in the local scheme, $55$ telematics features are then extracted from the anomaly scores vector according to Subsection~\ref{subsection:anomalyscores}. After computing all $55$ two-way interactions between the $11$ telematics features, five-fold cross-validation AUC is computed using a non-penalized logistic regression. The tuning results for the global scheme are shown in Figure~\ref{fig:tune_global_methods}.
It turns out that the optimal hyperparameter values found are $k = 50$ for LOF and $b = 400$ for IF, with respective AUCs of $0.5318$ and $0.5398$. These are the hyperparameter values that will be used from now on in the global scheme. The optimal hyperparameter for each global anomaly detection method and its corresponding AUC value are given in Table~\ref{tab:tuning_anomaly}.
\begin{table}[ht]
    \centering
    \begin{tabular}{l c c c c c}
        \toprule
        & \multicolumn{4}{c}{Hyperparameter value} & \\
        \cmidrule(l){2-5}
        \textbf{Anomaly detection algorithm} & $k_{frac}$ & $b_{frac}$ & $k$ & $b$ & AUC \\
        \midrule
        Local Mahalanobis & -- & -- & -- & -- & $0.5200^{(0.010079)}$\\
        Local LOF & $0.35$ & -- & -- & -- & $0.5450^{(0.009192)}$\\
        Local IF & -- & $0.85$ & -- & -- & $0.5320^{(0.009704)}$\\
        \cmidrule(l){1-6}
        Global Mahalanobis & -- & -- & -- & -- & $0.5481^{(0.006750)}$\\
        Global LOF & -- & -- & $50$ & -- & $0.5318^{(0.010498)}$\\
        Global IF & -- & -- & -- & $400$ & $0.5398^{(0.014215)}$\\
        \bottomrule 
    \end{tabular}
    \caption{Best hyperparameter value for each anomaly detection method and the resulting five-fold cross-validation AUC. Standard deviation is displayed as a superscript.}
    \label{tab:tuning_anomaly}
\end{table}

\subsection{Elastic-Net Logistic Regression}

Once LOF and IF are tuned for both local and global schemes, they, as well as Mahalanobis' method, are each employed to derive both a routine and a peculiarity profile for each vehicle, from which telematics features are extracted as quantiles according to Table~\ref{tab:preproc_ano_scores}. These telematics features, along with traditional risk factors (TRF) and distance driven, are then used together in an elastic-net logistic regression model. We have in total six different anomaly scores for each trip, six sets of telematics features are extracted. The classification performance of the six resulting models is compared to a baseline model that does not use anomaly detection at all, namely an elastic-net model that only uses the ten traditional risk factors and the distance driven as covariates. Remember that the elastic-net penalty has two hyperparameters, namely $\lambda$, which controls the severity of the penalty and $\alpha$, which represents the fraction of the lasso-type penalty to be used. Tuning is done separately for each of the seven models using grid search as the hyperparameter space searching strategy and AUC as the performance metric. For the $\lambda$ hyperparameter, a grid of $50$ log-uniformly distributed values between $10^{-10}$ and $1$, namely $\mathcal{G}_\lambda = \{10^{\frac{10}{49}i}\}_{i=0}^{49}$, is used. On the other hand, a coarse grid of five uniformly distributed values between $0$ and $1$ is considered for the mixing parameter $\alpha$, namely $\mathcal{G}_\alpha = \{0, 0.25, 0.5, 0.75, 1\}$. For each of the seven models, cross-validation is employed on the training set to find the best combination of hyperparameters. For a given elastic-net model, five-fold cross-validation AUC is thus computed for each of the $|\mathcal{G}_\lambda|\times |\mathcal{G}_\alpha| = 250$ hyperparameter combinations. The values for $\lambda$ and $\alpha$ that lead to the best cross-validation performance are thereafter chosen as the optimal pair. The optimal pair found for each of the seven models and the resulting AUC value are shown in Table~\ref{tab:tuning_results}.
\begin{table}[ht]
    \centering
    \begin{tabular}{l c c c}
        \toprule
        & \multicolumn{2}{c}{Optimal value} & \\
        \cmidrule(l){2-3}
        \textbf{Feature set used} & $\lambda$ & $\alpha$ & AUC \\
        \midrule
        TRF + Distance driven (baseline) & $2.33\times 10^{-2}$ & $1$ & $0.6096^{(0.01300)}$\\
        \cmidrule(l){1-4}
        TRF + Distance driven + Local Mahalanobis & $2.33\times 10^{-2}$ & $1$ & $0.6095^{(0.01289)}$\\
        TRF + Distance driven + Local LOF & $2.22\times 10^{-3}$ & $0.5$ & $0.6112^{(0.01392)}$\\
        TRF + Distance driven + Local IF & $2.33\times 10^{-2}$ & $1$ & $0.6096^{(0.01299)}$\\
        \cmidrule(l){1-4}
        TRF + Distance driven + Global Mahalanobis & $9.10\times 10^{-3}$ & $1$ & $0.6149^{(0.01550)}$\\
        TRF + Distance driven + Global LOF & $3.56\times 10^{-3}$ & $1$ & $0.6124^{(0.01410)}$\\
        TRF + Distance driven + Global IF & $9.10\times 10^{-3}$ & $0$ & $0.6126^{(0.01498)}$\\
        \bottomrule 
    \end{tabular}
    \caption{Hyperparameter tuning results for the seven elastic-net models.}
    \label{tab:tuning_results}
\end{table}

\section{Analyses}\label{sec:Analyses}

\subsection{Routine and peculiarity profiles visualization}\label{ssec:profile_viz}

The anomaly detection algorithms have been tuned in Subsection~\ref{ssec:Anomaly Detection Techniques} and are now ready to be applied to the processed telematics dataset in order to derive both a routine and a peculiarity profile for each vehicle, which we do. Ideally, we would like these profile vectors to help us differentiate between routine and non-routine vehicles, as well as between peculiar and non-peculiar ones. By inspecting the distribution of the trip attributes for each vehicle separately, we select two vehicle pairs: a routine/non-routine one and a peculiar/non-peculiar one. The former is shown in Figure~\ref{fig:routinerie} of Appendix~\ref{app:A} and it is clear that the first vehicle (Figure~\ref{fig:rout}) is more routine than the second (Figure~\ref{fig:non_rout}) in terms of the six trip attributes. Indeed, the former tends to make trips of about $13$ km concentrated around $11$:$00$ a.m. and $9$:$00$ p.m., while the latter makes trips of various distances at almost any time of the day. The attribute distributions of the selected peculiar/non-peculiar pair are presented in Figure~\ref{fig:peculiarite} of Appendix~\ref{app:A}, where the black density line corresponds to the global distribution of the attributes, i.e., the attributes' distributions on the whole telematics dataset. We have deemed the first vehicle peculiar because the distribution of its attributes does not match the global distribution very well, unlike the second vehicle, which is deemed non-peculiar.

\begin{figure}
  \centering
  \subcaptionbox{Routine profile for the routine/non-routine vehicle pair.\label{fig:plot_rout}}
    {\includegraphics[width = \textwidth]{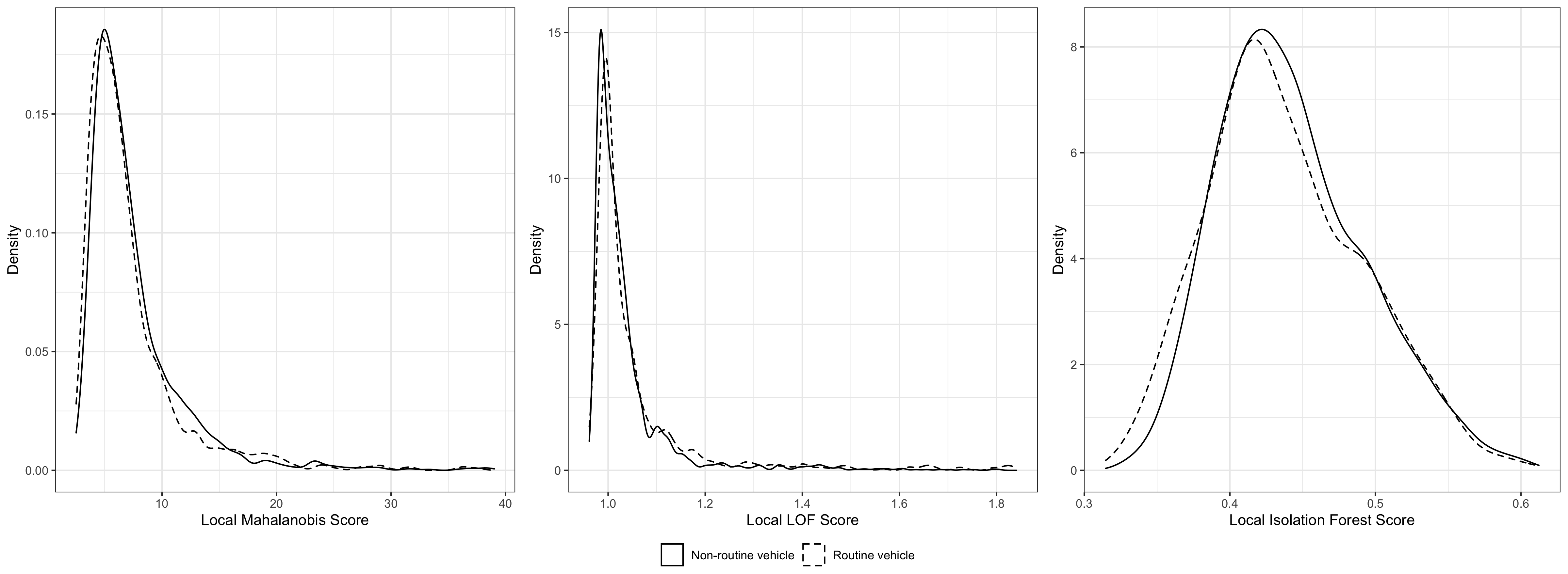}}
  \subcaptionbox{Peculiarity profile for the peculiar/non-peculiar vehicle pair.\label{fig:plot_pec}}
    {\includegraphics[width = \textwidth]{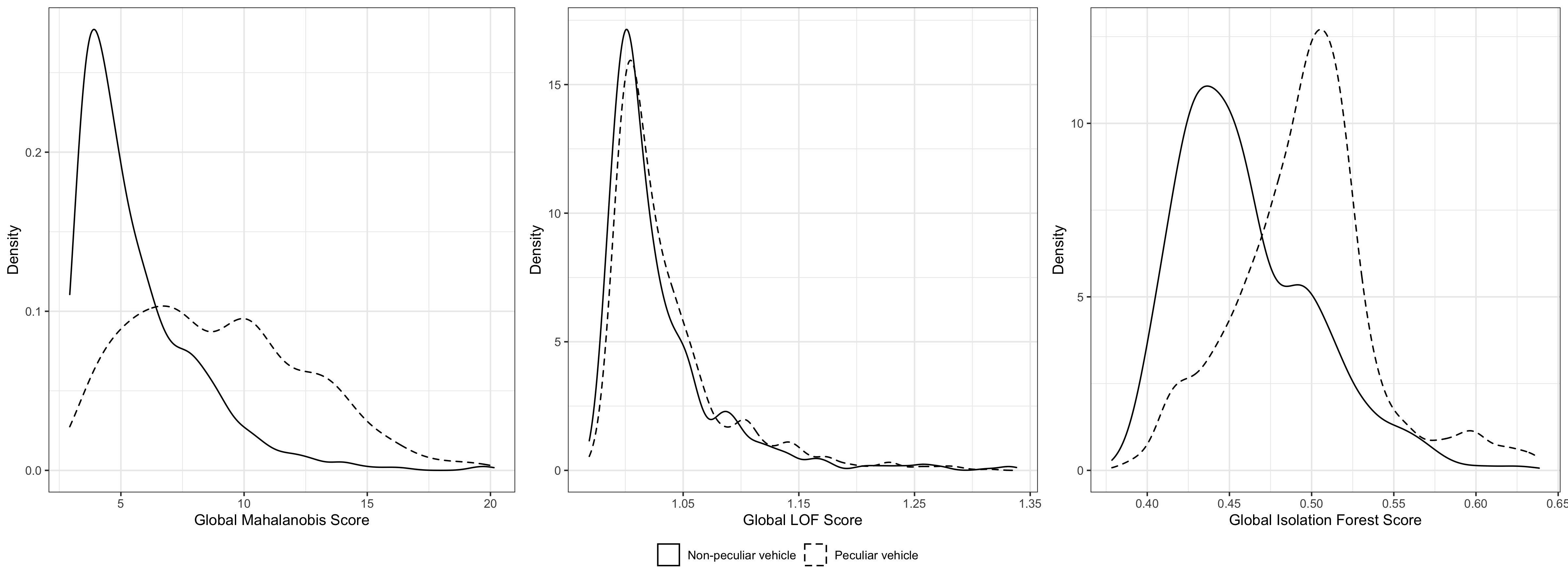}}
  \caption{Comparison of routine and peculiarity profiles for the routine/non-routine and the peculiar/non-peculiar vehicle pair.}\label{fig:profile_comparison}
\end{figure}

In Figure~\ref{fig:profile_comparison}, we compare the density of the routine and peculiarity profiles derived using each of the three anomaly detection algorithms for both the routine/non-routine and the peculiar/non-peculiar vehicle pairs. Looking at Figure~\ref{fig:plot_rout}, it is clear that anomaly detection algorithms applied locally fail to capture how routine a vehicle is. Indeed, the distribution of local anomaly scores for the selected routine vehicle coincides almost perfectly with that of the non-routine one, indicating that local anomaly scores computed with Mahalanobis' method, LOF and Isolation Forest all fail to distinguish a routine vehicle from a non-routine one. Conversely, global Mahalanobis and Isolation Forest scores seem to capture the peculiarity profile of vehicles well, unlike global LOF scores. Indeed, one sees that the peculiar vehicle has a distribution of global scores much further to the right of the real numbers axis than the non-peculiar one, indicating more unusual trips. As a result, we expect the global anomaly scores from Mahalanobis' method and the Isolation Forest to improve the ability to discriminate claimant from non-claimant vehicles, while local scores are expected to be of very little help.

\subsection{Classification results}

Elastic-net models, which were tuned in the previous section, are now ready to be trained on the full training set, which we do. They are then scored on the testing set, which has never been used before and is therefore totally unknown to the seven trained models. Classification performance is then measured with four metrics, namely AUC, accuracy, sensitivity and specificity. Accuracy is simply the number of correctly classified observations divided by the total number of observations. Sensitivity measures the ability of the model to correctly detect positive cases, i.e., claimants, while specificity measures the ability to correctly detect negative cases, i.e., non-claimants. The threshold used for the latter three metrics is $0.5$, which means the hard prediction is $0$ if the estimated probability of claiming is less than $0.5$ and is $1$ otherwise. In Table~\ref{tab:testing_results}, the improvement of the six models that use anomaly detection-based features over the baseline model is displayed, as well as the baseline model's absolute performance.
\begin{table}[ht]
    \centering
    \begin{adjustbox}{max width = \textwidth}
    \begin{tabular}{l c c c c}
        \toprule
        \textbf{Anomaly detection algorithm} & AUC & Accuracy & Sensitivity & Specificity\\
        \midrule
        TRF + Distance driven (baseline) & $0.5948$ & $0.5634$ & $0.5862$ & $0.5407$ \\
        \cmidrule(l){1-5}
        TRF + Distance driven + Local Mahalanobis & $0.0001$ & $0$ & $0$ & $0$ \\
        TRF + Distance driven + Local LOF & $-0.0013$ & $-0.0013$ & $-0.0193$ & $0.0165$ \\
        TRF + Distance driven + Local IF & $0$ & $0$ & $0$ & $0$ \\
        \cmidrule(l){1-5}
        TRF + Distance driven + Global Mahalanobis & $0.0184$ & $0.0214$ & $0.0110$ & $0.0317$ \\
        TRF + Distance driven + Global LOF & $0.0006$ & $0.0042$ & $0.0028$ & $0.0055$ \\
        TRF + Distance driven + Global IF & $0.0117$ & $0.0063$ & $-0.0372$ & $0.0496$ \\
        \bottomrule 
    \end{tabular}
    \end{adjustbox}
    \caption{Improvement of the six elastic-net models using anomaly detection over the baseline model and the baseline model's performance.}
    \label{tab:testing_results}
\end{table}
The latter achieves an AUC of $0.5948$, which is fine considering the small amount of data. In point of fact, the elastic-net models are tuned and trained on the training set, which has $3384$ rows, and are scored on the testing set, which has $1450$ rows. Moreover, AUC values are usually fairly low in claim classification due to the inherent randomness of the problem. 

Clearly, features extracted from global anomaly scores, namely features extracted from the peculiarity profiles, are better candidates for classification than those extracted from local scores because they consistently yield superior results. In fact, the features extracted from local anomaly scores are deemed completely useless by the classification model because their inclusion does not improve (or even worsens) the performance metrics. This means that the routine profile we derived does not influence the risk of a claim. Conversely, global anomaly scores-based features nearly always improve classification metrics, regardless of whether Mahalanobis, LOF or IF is used to derive the anomaly scores. Global Mahalanobis performs best, with an improvement in AUC, accuracy, sensitivity and specificity of $0.0184$, $0.0214$, $0.0110$ and $0.0317$ over the baseline model, respectively. LOF is the worst among global algorithms overall, showing only a slight improvement over the baseline model. We successfully extracted useful features from the global anomaly scores, indicating that the peculiarity profile of a vehicle impacts the probability of claiming. This is in line with the analysis of the routine and peculiarity profiles performed on the two selected pairs of vehicles in Subsection~\ref{ssec:profile_viz}.

\begin{figure}
    \centering
    \includegraphics[width = \textwidth]{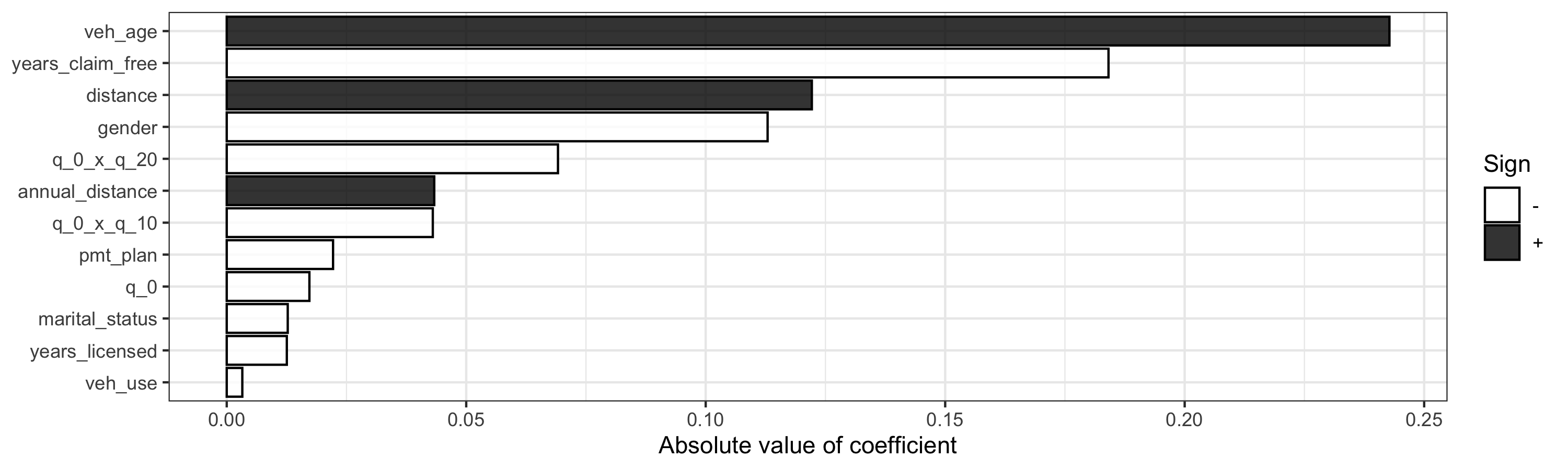}
    \caption{Elastic-net logistic regression non-zero coefficients obtained using features derived from global Mahalanobis anomay scores.}
    \label{fig:coefs_global_maha}
\end{figure}

We shall now examine how the peculiarity profile of a vehicle influences the claim risk. To this end, we analyze the estimated coefficients of our best model, namely the one using features derived from global Mahalanobis scores. The sign and the absolute value of each estimated coefficient indicates how its related feature influences the probability of claiming. Furthermore, because features have been standardized prior to training the model, the absolute value of a coefficient can be interpreted as a measure of the importance of the underlying feature. In Figure~\ref{fig:coefs_global_maha}, the sign and absolute value of every non-zero coefficient is displayed.
Interestingly, out of the $77$ features given as input to the model, only $12$ are assigned a non-zero coefficient, which means the model deemed the remaining $65$ features irrelevant for the classification task. The $12$ features selected by the model include eight traditional risk factors (\texttt{veh\_age}, \texttt{years\_claim\_free}, \texttt{gender}, \texttt{annual\_distance}, \texttt{pmt\_plan}, \texttt{marital\_status}, \texttt{years\_licensed} and \texttt{veh\_use}), distance driven (\texttt{distance}) and three features extracted from global Mahalanobis scores (\texttt{q0\_x\_q20}, \texttt{q0\_x\_q10} and \texttt{q0}). Interestingly enough, among the $11$ extracted telematics features (\texttt{q0}, \texttt{q10}, \dots, \texttt{q90}, \texttt{q100}) and their $55$ two-fold interactions (\texttt{q0\_x\_q10}, \texttt{q0\_x\_q20}, \dots, \texttt{q80\_x\_q90}, \texttt{q90\_x\_q100}), only features involving the $0^\text{th}$ percentile (i.e. the minimum of the anomaly score vector) and low percentiles are kept by the model, namely the $0^\text{th}$ percentile itself (\texttt{q0}), the interaction between the $0^\text{th}$ and $10^\text{th}$ percentiles (\texttt{q0\_x\_q10}) and the interaction between the $0^\text{th}$ and $20^\text{th}$ percentile (\texttt{q0\_x\_q20}). This indicates that lower global anomaly scores for a vehicle are predictive of the occurrence of a claim. Moreover, the coefficients of the abovementioned three anomaly score-based features are negative, which means, \emph{ceteris paribus}, that the lower the percentile of the scores of a vehicle, the more likely it is to claim. This may seem counterintuitive because one might think that more anomalous trips must be more risky. However, when analyzing correlations between global Mahalanobis score and the eight trip attributes in Table~\ref{tab:corr_trip_attributes}, one notices that the anomaly score is more closely linked to \texttt{duration}, \texttt{distance}, \texttt{avg\_speed} and \texttt{max\_speed} than attributes related to time of day and time of week. 
\begin{table}[ht]
    \centering
    \begin{tabular}{l c}
        \toprule
        \textbf{Trip attribute} & \textbf{Spearman correlation}\\
        \midrule
        \texttt{duration} & $0.62$\\
        \texttt{distance} & $0.79$\\
        \texttt{avg\_speed} & $0.35$\\
        \texttt{max\_speed} & $0.28$\\
        \texttt{time\_of\_day\_sin} & $0.09$\\
        \texttt{time\_of\_day\_cos} & $0.09$\\
        \texttt{time\_of\_week\_sin} & $-0.02$\\
        \texttt{time\_of\_week\_cos} & $0.02$\\
        \bottomrule 
    \end{tabular}
    \caption{Spearman correlation of the eight trip attributes with the global Mahalanobis anomaly score.}
    \label{tab:corr_trip_attributes}
\end{table}
This is because the four axes \texttt{time\_of\_day\_sin}, \texttt{time\_of\_day\_cos}, \texttt{time\_of\_week\_sin} and  \texttt{time\_of\_week\_cos} have fewer anomalous values than the axes \texttt{duration}, \texttt{distance}, \texttt{avg\_speed} and \texttt{max\_speed}. Therefore, the anomalies are mostly determined using the first four attributes in Table~\ref{tab:corr_trip_attributes}. Given that the four attributes \texttt{duration}, \texttt{distance}, \texttt{avg\_speed} and \texttt{max\_speed} have a strong positive correlation with the anomaly score, this means that trips that are identified as anomalous (i.e., trips with a high anomaly score) are also trips with high values for the four abovementioned attributes. Naturally, trips with high values for these four attributes are most likely to have been taken on highways, which are usually safer than other types of roads. This would explain the negative coefficients associated with the global Mahalanobis scores-based features \texttt{q0}, \texttt{q0\_x\_q10} and \texttt{q0\_x\_q20}.

\section{Conclusions}\label{sec:Conclusions}

In this study, a novel method that allows the incorporation of telematics data into a claim classification model has been developed. The procedure, which utilizes anomaly detection algorithms, automatically extracts features from summarized information about trips made by insured vehicles. While most studies create telematics features by hand and subjectively, which may favor or disfavor certain insureds, the approach developed in this paper has the benefit of being objective. Although the method has been tested in the context of claim classification, it could easily be applied to the context of frequency or even severity modeling, because these are supervised learning problems as well. We found that anomaly scores computed globally can distinguish peculiar from non-peculiar vehicles and using a thorough machine learning methodology, we showed that these scores convey important information about the claiming risk of a vehicle. Indeed, features extracted from them allow the improvement of classification performance when added to the baseline model, especially when Mahalanobis' method is at play. We also found out that lower quantiles of the peculiarity profile vector are the most predictive of the claiming risk, and that these quantiles are negatively correlated with the probability of claiming. It seems that locally calculated anomaly scores, which is to say the routine profiles, do not help to improve classification, and this is probably due to the fact that such anomaly scores are unable to discern routine from non-routine vehicles.

In our method, each instance, or vehicle, is described with a distribution of local and global anomaly scores. From these distributions, quantiles are extracted and used as features in a penalized logistic regression model. While this type of model has a good performance/interpretability ratio, it is not the most appropriate for detecting complex interactions between features, which we think may be useful to properly characterize the anomaly scores' distribution and thus improve prediction. Such models also deal poorly with non-linear relationships between features and response. In fact, the current model only allows for linear links between features and log-odds, and uses the extracted telematics features as is, without trying to make them interact. For future research, we believe it would be wise to apply a neural network on the quantiles extracted from the routine and peculiarity profiles. Indeed, neural networks are known to automatically learn useful features from raw data. This way, instead of being arbitrary, the feature extraction process would be guided by the feedback of a loss function, which would probably enhance prediction. Alternatively, one could apply a neural network directly to the routine and peculiarity profile vectors. Furthermore, we still believe that the trips' degree of routine could possibly help to better classify the vehicles, but that the anomaly detection algorithms as implemented had a hard time capturing this degree of routine. Thus, for future research, one could try to find a way to better capture vehicles' routine profiles.


\clearpage

\section*{Acknowledgement}

The authors gratefully acknowledge The Co-operators for both financial support and for providing the data used in this paper through the Co-operators Chair in Actuarial Risk Analysis.

\section*{Funding}

The authors thank The Co-operators and the Natural Sciences and Engineering Research Council of Canada for funding.

\bibliographystyle{apalike}  
\bibliography{main}


\clearpage
\appendix

\section{Routine/non-routine and peculiar/non-peculiar vehicle pairs}\label{app:A}

\begin{figure}[ht]
  \centering
  \subcaptionbox{Routine selected vehicle\label{fig:rout}}{\includegraphics[width = \textwidth]{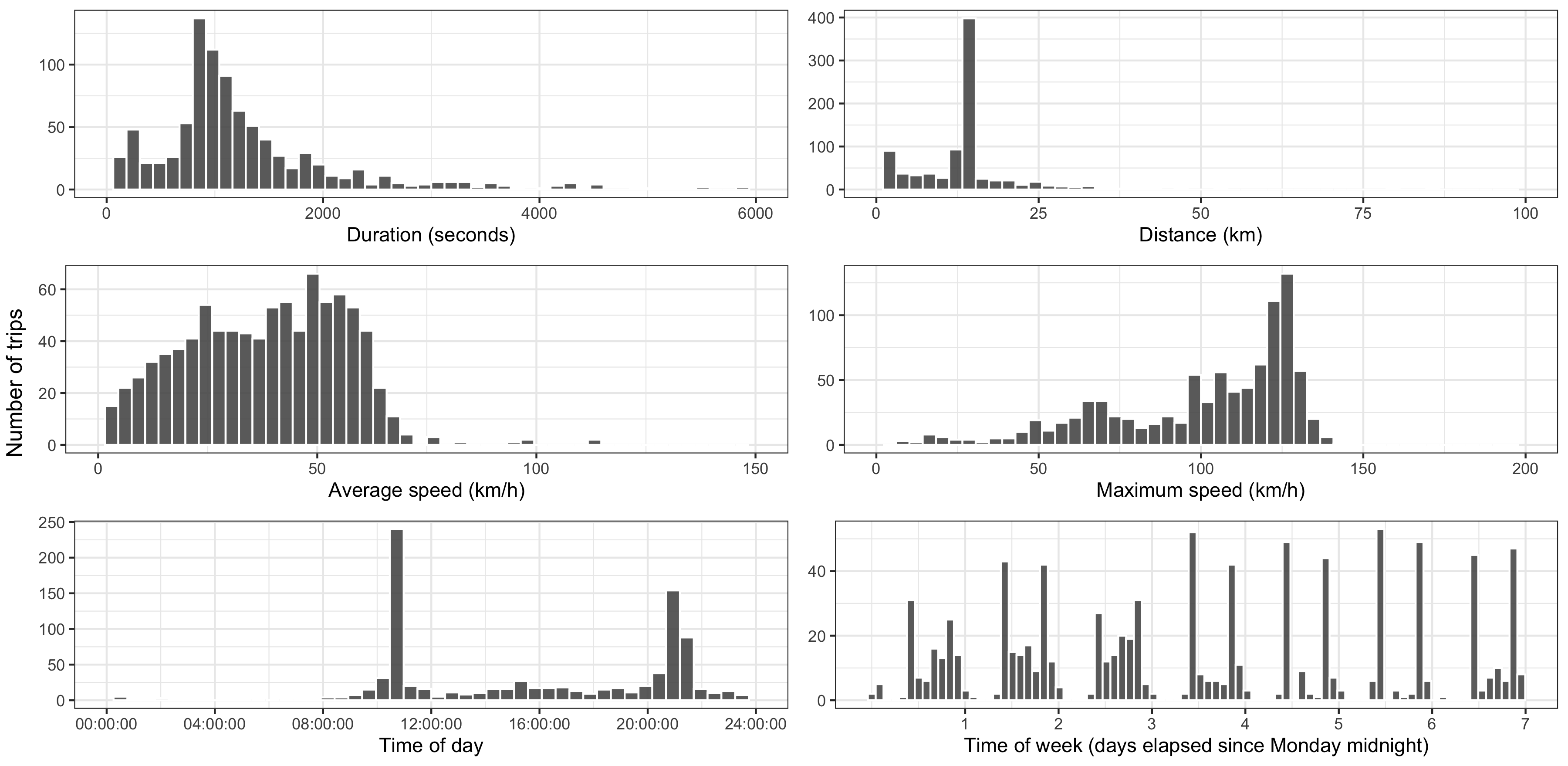}}
  \subcaptionbox{Non-routine selected vehicle\label{fig:non_rout}}{\includegraphics[width = \textwidth]{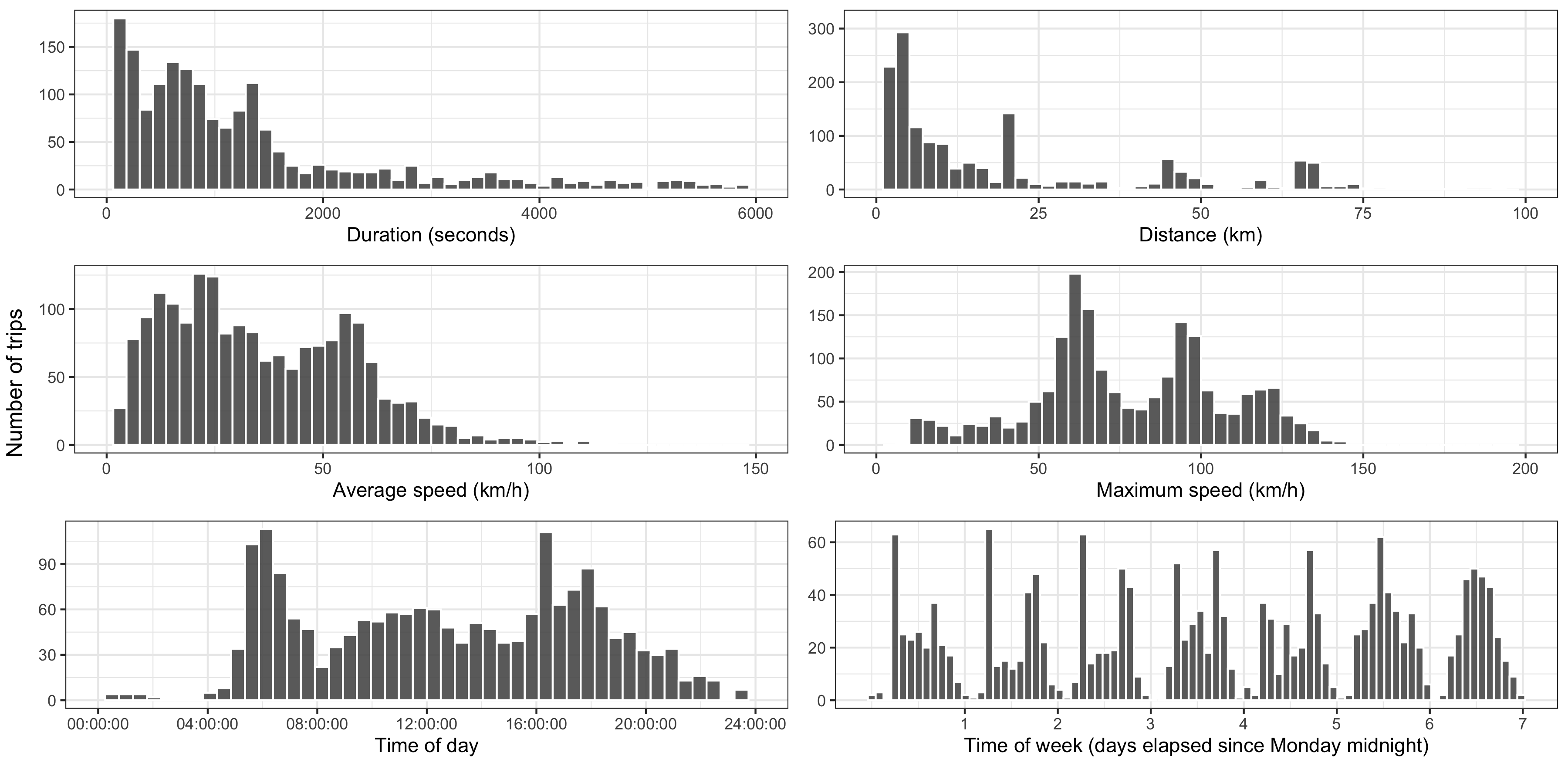}}
  \caption{Histograms of trip attributes for the routine/non-routine pair.}\label{fig:routinerie}
\end{figure}

\begin{figure}[ht]
  \centering
  \subcaptionbox{Peculiar selected vehicle\label{fig:pec}}{\includegraphics[width = \textwidth]{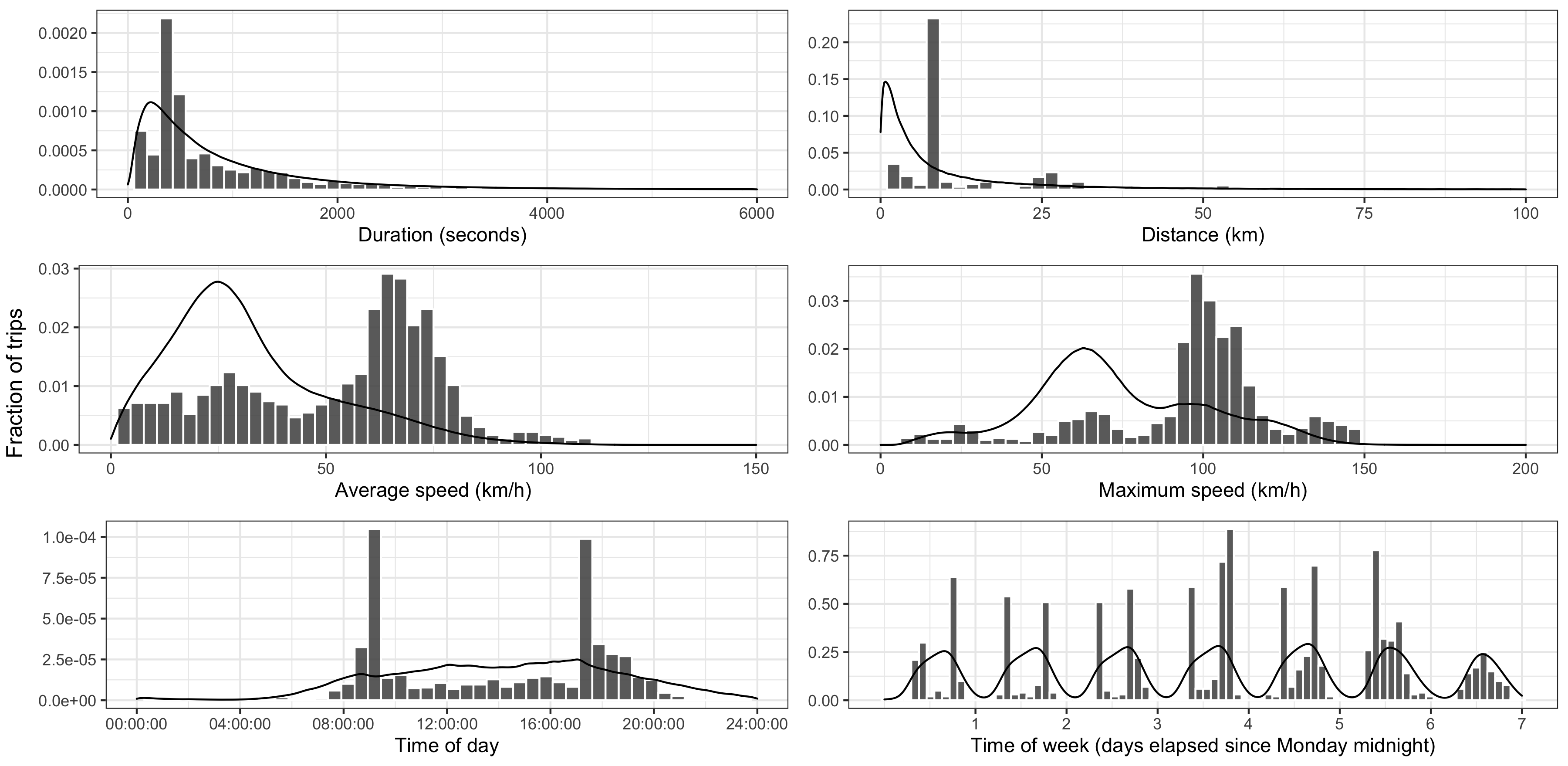}}
  \subcaptionbox{Non-peculiar selected vehicle\label{fig:non_pec}}{\includegraphics[width = \textwidth]{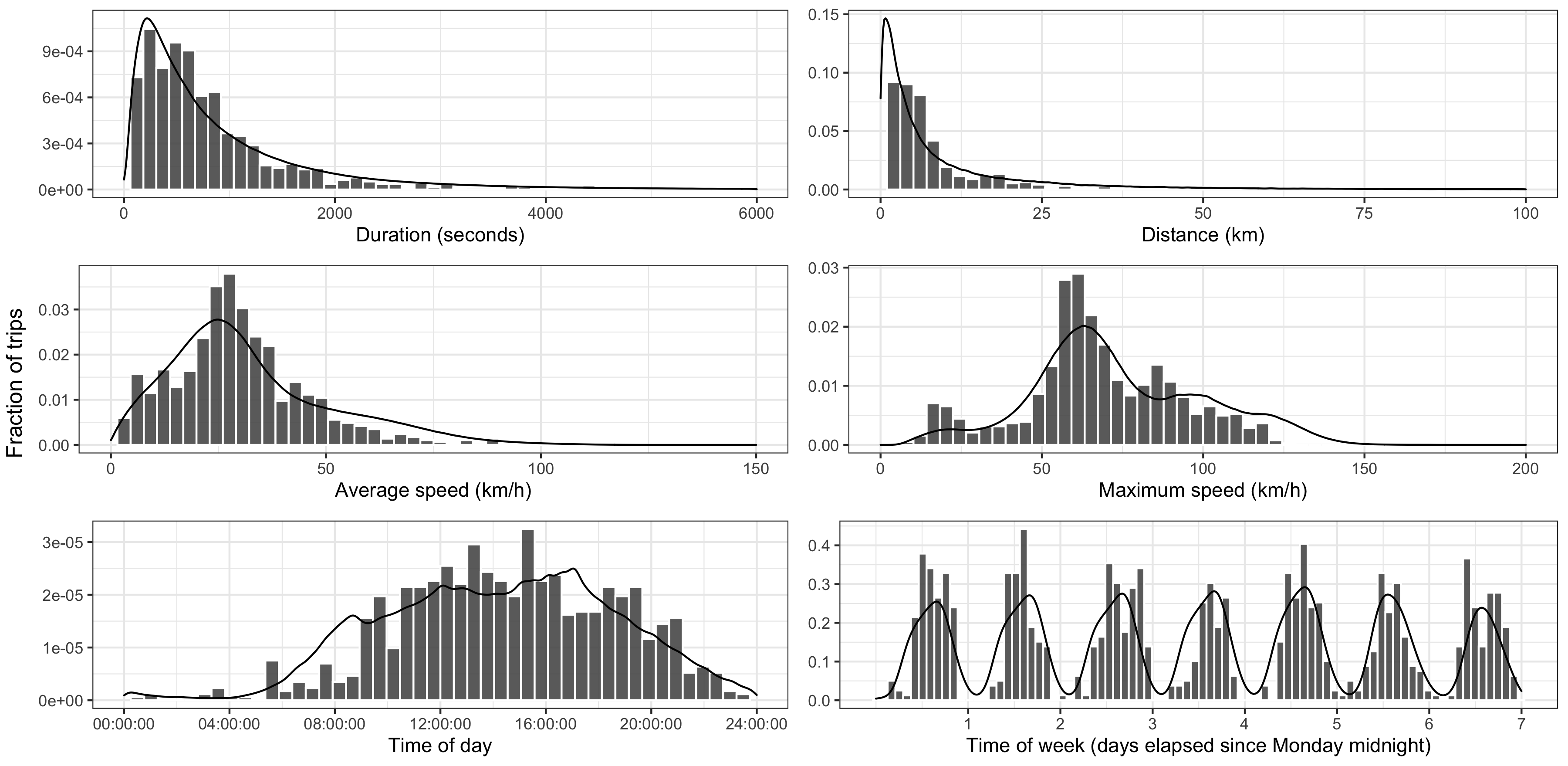}}
  \caption{Histograms of trip attributes for the peculiar/non-peculiar pair. The black line shows the densities of the attributes for the whole telematics dataset.}\label{fig:peculiarite}
\end{figure}


\end{document}